\documentclass[10pt,twocolumn,letterpaper]{article}
\usepackage{iccv}
\usepackage{times}
\usepackage{epsfig}
\usepackage{graphicx}
\usepackage{amsmath}
\usepackage{amssymb}
\usepackage{times}  
\usepackage[hyphens]{url}  
\usepackage{graphicx} 
\usepackage{caption}
\usepackage{subcaption}
\usepackage{multirow}
\usepackage{physics}
\usepackage[accsupp]{axessibility}

\usepackage{algorithm}
\usepackage{algorithmic}

\usepackage{newfloat}
\usepackage{listings}
\usepackage{color}

\usepackage[pagebackref=true,breaklinks=true,colorlinks,bookmarks=false]{hyperref}

\iccvfinalcopy 

\def\iccvPaperID{7499} 

\ificcvfinal\fi

\begin{document}

\title{Point-TTA: Test-Time Adaptation for Point Cloud Registration \\ Using Multitask Meta-Auxiliary Learning}

\author{Ahmed Hatem\\
University of Manitoba\\
{\tt\small hatema@myumanitoba.ca}
\and
Yiming Qian\\
University of Manitoba\\
{\tt\small yiming.qian@umanitoba.ca}
\and
Yang Wang\\
Concordia University\\
{\tt\small yang.wang@concordia.ca}}

\maketitle
\ificcvfinal\thispagestyle{empty}\fi

\begin{abstract}
We present Point-TTA, a novel test-time adaptation framework for point cloud registration (PCR) that improves the generalization and the performance of registration models. While learning-based approaches have achieved impressive progress, generalization to unknown testing environments remains a major challenge due to the variations in 3D scans. Existing methods typically train a generic model and the same trained model is applied on each instance during testing. This could be sub-optimal since it is difficult for the same model to handle all the variations during testing. In this paper, we propose a test-time adaptation approach for PCR. Our model can adapt to unseen distributions at test-time without requiring any prior knowledge of the test data. Concretely, we design three self-supervised auxiliary tasks that are optimized jointly with the primary PCR task. Given a test instance, we adapt our model using these auxiliary tasks and the updated model is used to perform the inference. During training, our model is trained using a meta-auxiliary learning approach, such that the adapted model via auxiliary tasks improves the accuracy of the primary task. Experimental results demonstrate the effectiveness of our approach in improving generalization of point cloud registration and outperforming other state-of-the-art approaches.

\end{abstract}

\section{Introduction}
Given a pair of overlapping 3D point clouds, the goal of point cloud registration (PCR) is to estimate the 3D transformation that aligns these point clouds. PCR plays a vital role in a variety of applications, such as autonomous driving~\cite{3Dauto},  augmented reality~\cite{ar_survey}, and robotics~\cite{robotics_survey}. Standard PCR approaches start with extracting the pointwise features. These features are matched to establish the correspondences between points in the pair. Then an outlier rejection module is used to remove outliers since the correspondences obtained by feature matching are not completely reliable. Finally, the correspondences are used to estimate the optimal 3D rotation and translation to align the two point cloud fragments. Traditional approaches establish correspondences by matching hand-crafted features and leveraging robust iterative sampling strategies such as RANSAC \cite{ransac} for model estimation. However, these approaches take a long time to converge and their accuracy drops in the presence of high outliers. 

To address the limitation of traditional approaches, most recent PCR methods use learning-based approaches instead of handcrafted features. These approaches first learn a model on a labeled dataset. Then the model is fixed when evaluating on unseen test data. The single set of model parameters may not be optimal for different test environments captured with different 3D scanners due to the domain shift. Our work is inspired by the success of test-time adaption (TTA)~\cite{ttt} in image classification, where an auxiliary task is used to update the model parameters at inference time to learn feature representation specifically for a test instance. In this paper, we introduce a TTA approach for point cloud registration that adapts to new distributions at test-time. 
Unlike existing 3D point cloud domain adaptation approaches \cite{adaptPoint2,adaptPoint3,adaptPoint4,adaptPoint5,adaptPoint6}, our method does not require any prior knowledge about the test data distributions. 
We adapt the model parameters in an instance-specific manner during inference and obtain a different set of network parameters for each different instance. This allows our model to better capture the uniqueness of each test instance and thus generalize better to unseen data. More importantly, our TTA formulation is a \emph{generic} framework that can be applied in a plug-and-play manner to boost standard PCR pipelines. 

Auxiliary learning has been shown to be effective to improve a predefined primary task and is used in multiple 2D computer vision tasks \cite{maxl,depth_aux_2020_cvpr, ttt,deblurring_2021_cvpr}. Recently, there have been some attempts at using auxiliary tasks for improving the representation learning of 3D point clouds \cite{adaptPoint,3D_byol,3D_reconst,sslAdapt3}. However, using auxiliary tasks for test-time adaptation is still largely unexplored for 3D point cloud data.  In this work, we introduce three self-supervised auxiliary tasks: point cloud reconstruction, feature learning, and correspondence classification. These auxiliary tasks do not require any extra supervision. 

Some recent work \cite{deblurring_2021_cvpr} has shown that naively training the primary task and the auxiliary task together may not be optimal, since the model may be biased toward improving the auxiliary task rather than the primary task. Following \cite{deblurring_2021_cvpr}, we use a meta auxiliary learning framework based on the model-agnostic meta learning (MAML) \cite{maml}. Each point cloud pair acts as a task in MAML. We update the model for each task using the auxiliary loss provided by the proposed three auxiliary tasks. The updated model is then used to perform the primary task. The model parameters are learned in such a way that the updated model using the auxiliary tasks improve the performance of the primary registration task. Our key contributions are summarized as follows:
\begin{itemize}
  \item We propose a test-time adaptation approach for point cloud registration. To the best of our knowledge, this is the first work to apply test-time adaptation for 3D point cloud registration.
  \item  We design three self-supervised auxiliary tasks to effectively extract useful features from test instances and adapt the model to unseen test distribution to improve generalization. A meta auxiliary learning paradigm is used to learn the model parameters, such that adapting the model parameter via the auxiliary tasks during testing improves the performance of the primary task. 
  \item We perform extensive experiments on 3Dmatch \cite{3Dmatch} and KITTI \cite{kitti} benchmarks to show the effectiveness of our approach in improving point cloud registration performance and achieving superior results.
\end{itemize}
 
\section{Related Work}
We review two lines of related work in point cloud registration and meta-auxiliary learning.

\paragraph{Point Cloud Registration.} Most traditional PCR approaches consist of two modules: feature-based correspondence matching and outlier filtering.

Feature descriptors have been proposed to effectively extract the local and global features of point clouds, which are used to match correspondences in the feature space. Traditional methods use hand-crafted features such as spatial features histogram \cite{spHistogram1,spHistogram2,2004Hist}, or geometric features histogram \cite{geoHist,fpfh}. Recently, learning-based approaches have been proposed to learn 3D feature descriptors including fully convolution methods \cite{perfectMatch_2019_CVPR,fcgf}, keypoint detection methods \cite{D3Feat_2020_CVPR,3dfeat,usip}, and coarse-to-fine methods \cite{CoFiNet,geoTF}. 

The correspondences obtained from feature matching often include many outliers that must be filtered out for robust point cloud registration. Many traditional approaches \cite{ransac,fgr,teaser,sc2} have been proposed for robust outlier filtering of correspondences. RANSAC \cite{ransac} is the most popular method, where a set of correspondences are iteratively sampled to filter outliers. RANSAC variants \cite{ransac2,ransac3,GCransac} have been introduced to provide new sampling strategies for fast convergence. However, these methods still have slow convergence rate and low performance in the presence of high outliers. Other methods use robust cost functions that are more effective with high outlier ratio. FGR \cite{fgr} uses the Geman-McClure cost function and TEASER \cite{teaser} uses the truncated least squares cost function for robust point cloud registration.

In recent years, deep learning techniques have been employed for outlier filtering. The 3D outlier filtering approaches \cite{3dreg,gmf,robust_2022,dgr,dhvr,pointDSC} follow similar ideas in 2D image matching \cite{2dcorr1,2dcorr2}, where outlier filtering is defined as an inlier classification problem. 3DRegNet \cite{3dreg} uses the 2D correspondence selection network \cite{2dcorr1} for 3D point clouds and added a regression module for rigid transformation. DGR \cite{dgr} proposes a fully convolutional network to better capture global features of correspondences and predict the inlier confidence of each correspondence. DHVR \cite{dhvr} leverages Hough voting in 6D transformation parameter space to identify the confidence of correspondences from Hough space to predict the final transformation. PointDSC \cite{pointDSC} uses the spatial consistency between inlier correspondences to better prune the outliers.

\paragraph{Auxiliary and Meta Learning.}
In auxiliary learning, an auxiliary task (often self-supervised) is defined to improve the performance and generalization of a target primary task. This differs from multi-task learning where the goal is to improve performance across all tasks \cite{maxl}. Auxiliary learning has been proven to be effective in multiple 2D image domain problems. \cite{ttt} uses the image rotation prediction as a self-supervised auxiliary task to improve image classification. \cite{deblurring_2021_cvpr} uses image reconstruction as the auxiliary task to improve the primary task of deblurring. Auxiliary learning has also been studied for 3D point clouds. \cite{3D_reconst} proposes an auxiliary task that reconstruct point clouds whose parts have been randomly displaced. \cite{3D_byol} adopts a contrastive auxiliary task \cite{byol_nips_2020} for better representation of 3D point clouds. Moreover, several studies have proposed self-supervised
tasks \cite{surveySSL, adaptPoint, sslAdapt3} to learn domain-invariant and useful representations of point clouds from unlabelled point cloud data. \cite{adaptPoint} presents a self-supervised task of reconstructing deformations to learn the underlying structures of 3D objects. \cite{sslAdapt3} introduces a pair of self-supervised tasks, including a scale prediction task and a 3D/2D projection reconstruction task to facilitate  global and local features learning across different domains.

\par MAML \cite{maml} is a widely used meta-learning algorithm, which has been successfully employed for many 2D image domain tasks \cite{metaResolution,metaResolution2,maxl,meta_depth,deblurring_2021_cvpr}. MAML learns the model parameters to fastly adapt to new tasks with few training samples and few gradient updates. \cite{metaResolution,metaResolution2} use MAML for super-resolution problem to facilitate adaptation to unseen images. \cite{maxl} proposes a meta-auxiliary framework (MAXL), in which an auxiliary label generator is trained to generate optimal labels to improve the generalization of the primary task. \cite{deblurring_2021_cvpr} uses a self-supervised auxiliary task in a meta learning framework for image deblurring. \cite{meta_depth} propose to employ meta-auxiliary learning along with test-time adaptation for the problem of future depth prediction in videos. 



\section{Approach}

\begin{figure*}
\centering
\includegraphics[width=\textwidth]{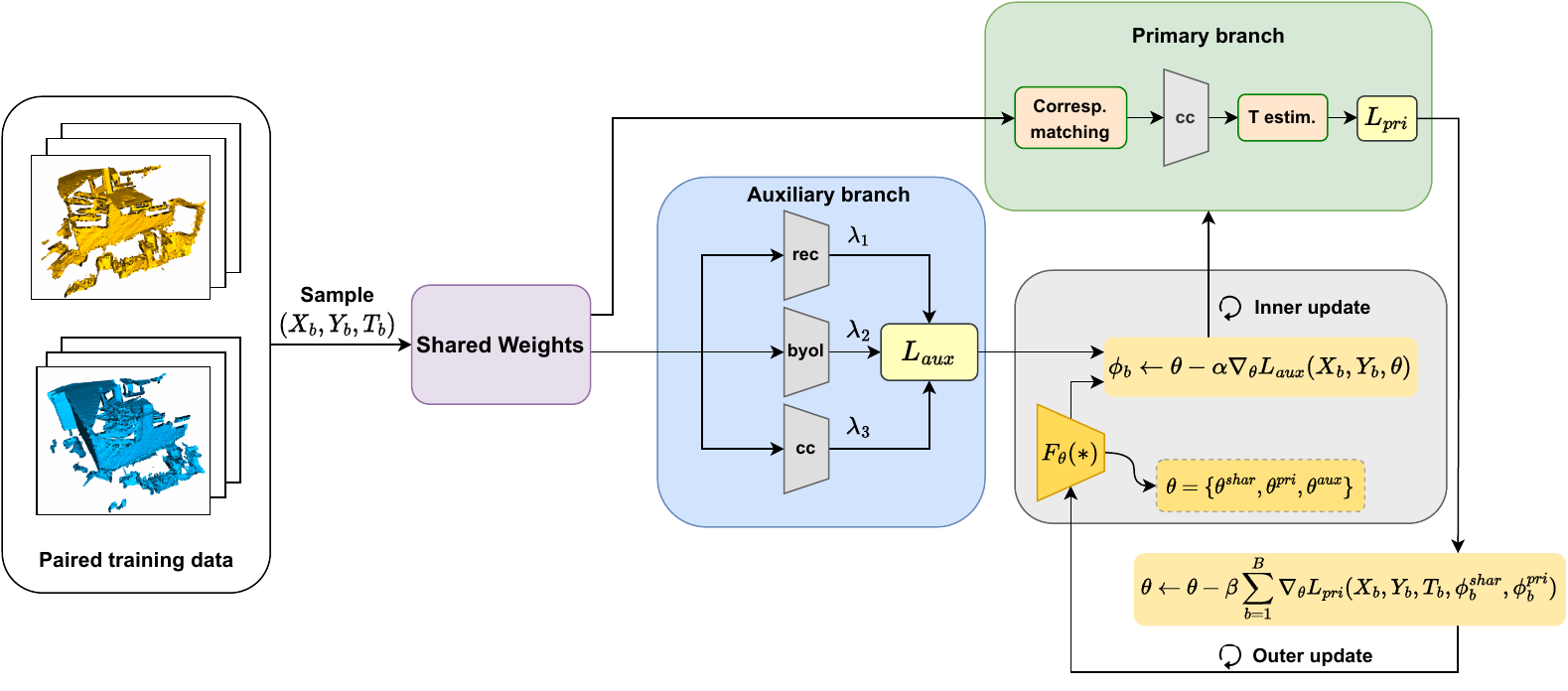}
   \caption{Overview of the proposed meta-auxiliary training framework. Given a pair of input point clouds during training, we first adapt the model by performing a small number of gradient updates using the auxiliary loss calculated via three auxiliary tasks including point cloud reconstruction, BYOL and correspondence classification. Then the adapted model is used to perform the primary registration task and is evaluated using the meta-objective. Finally, we update the model using the primary loss. }
   
\label{framework}
\end{figure*}

\par Given a pair of partially overlapping 3D point clouds $ X \in R^{M \times 3}$ with $M$ points and $Y \in R^{N \times 3} $ with $N$ points, our goal is to find an optimal 3D transformation $T$ between the two point clouds that accurately aligns them. Our goal is to learn a model $F_{\theta}(X,Y)\rightarrow T$ parameterized by $\theta$ that maps $(X,Y)$ to $T$. In this work, we propose a framework for point cloud registration that adapts the trained model parameters to each different input at test time, so that our model can improve generalization and performance of point cloud registration. The adaptation is achieved via self-supervised auxiliary tasks. 


\subsection{Auxiliary Tasks}\label{multi_aux}
We propose three different auxiliary tasks in our work. All these auxiliary tasks are self-supervised and do not require extra labels. So they can be used during test-time for adaptation.

\noindent{\bf Point Cloud Reconstruction.} Inspired by the success of using image reconstruction as the auxiliary task \cite{deblurring_2021_cvpr,depth_aux_2020_cvpr}, we propose to use 3D point cloud reconstruction as one of our self-supervised auxiliary tasks. Given a point cloud $P$, the features of the point cloud are extracted using the feature encoder. Then, a decoder is used to reconstruct the point cloud $P'$. Adapting the model parameters at test time using the reconstruction auxiliary loss enables the model to take advantage of the internal features of the test instance before performing the primary task. The reconstruction loss does not require any supervision, which makes it suitable for test-time adaptation. We use $L1$ reconstruction loss as follows:
\begin{equation}
 \ell_{rec} = ||P - P'||_1.
\end{equation}

\noindent{\bf Self-Supervised Feature Learning.} Self-supervised learning (SSL) is an active area of research. The main idea of SSL is to define some proxy self-supervised tasks to learn feature representations from data without manual annotations. We can use any existing SSL task as one of our auxiliary tasks. In our work, we adapt BYOL \cite{byol_nips_2020} as our self-supervised task. Different from contrastive learning, BYOL does not require negative samples. This makes it suitable for test-time adaptation. The model architecture of BYOL consists of two networks, namely the online network and the target network. Each network predicts a representation of an augmented view of the same point cloud. The idea is to train the online network to predict representations similar to the target network’s predictions, so that the representations of the two augmented views are closely similar.

The online network parameterized by $\theta$ consists of a feature encoder $f_{\theta}$, a feature projector $z_{\theta}$ and a predictor $p_{\theta}$. Similarly, the target network parameterized by $\xi$ has a feature encoder $f_{\xi}$ and a feature projector $z_{\xi}$ . The online network $\theta$ is trained based on the regression targets provided by the target network, while the target network $\xi$ is the exponential moving average of the online parameters $\theta$:
\begin{equation}
 \xi \leftarrow \tau\xi + (1-\tau) \theta,
\end{equation}
where $\tau \in [0, 1]$ is the target decay rate.

\par Given a 3D point cloud $P$, we perform augmentation to produce two augmented versions $P_{v}$ and $P_{v'}$. The point $P_{v}$ is passed to the online network to obtain the projection $z_{\theta} = g_{\theta}(P_{v})$ and $P_{v'}$ is passed to the target network to obtain the projection $z_{\xi} = g_{\xi}(P_{v'})$. Then we minimize the mean squared error between the normalized predictions $q_{\theta}(z_{\theta})$ and target projections $z_{\xi}$ as follows:
\begin{equation}\label{equ_byol}
 L_{\theta, \xi}  =  2 -  \frac{2 q_{\theta}(z_{\theta})^\top  z_{\xi}}{\Vert q_{\theta}(z_{\theta})\Vert^2 \Vert\norm{z_{\xi}\Vert^2}}.
\end{equation}

We define another symmetric loss $L'_{\theta,\xi}$ by similarly passing $P_{v'}$ to the online network and $P_{v}$ to the target network to compute $L'_{\theta,\xi}$. The final BYOL loss is defined as:  
\begin{equation}
\ell_{byol} =L_{\theta,\xi} + L'_{\theta,\xi}.
\end{equation}

\noindent{\bf Correspondence Classification.} We introduce an additional self-supervised auxiliary task designed specifically for PCR. Given a 3D point cloud $P$, we construct an augmented point cloud $P'$ using a randomly generated 3D transformation $T$ by sampling a random rotation along three axes within $[0^{\circ} .. 360^{\circ}]$ and a random translation within [0cm..60cm]. The sampled transformation $T$ is applied on each axis of point cloud to obtain $P'$. The feature encoder is used to extract the features of $P$ and $P'$. Then these two sets of points are matched in the feature space using nearest neighbors to obtain the correspondences.
 Using the same outlier rejection network architecture of the primary task, this auxiliary task is trained to predict whether a correspondence is an inlier or an outlier. Since the transformation $T$ of the point cloud is known, the ground-truth inlier correspondences $C$ are available and the auxiliary loss does not require any manual supervision. Similar to \cite{dgr,pointDSC}, the classification loss is defined as the binary cross entropy loss between the probability $p_{(i,j)}^i$ that a correspondence $C_{(i,j)}$ is an inlier and the ground-truth inliers $C$.  
 \begin{equation}
\ell_{cc} = \frac{1}{|M|} (\sum_{(i,j) \in C} \log{p_{(i,j)}^i}  + \sum \log{p_{(i,j)}^o )},
\end{equation}
where $p^o = 1 - p^i$.

\subsection{Model Architecture}
Our model architecture consists of a shared feature encoder and two branches for the primary and auxiliary tasks. The primary branch corresponds to the point cloud registration task. The auxiliary branch corresponds to three self-supervised auxiliary tasks defined in Section \ref{multi_aux}. We denote the model parameters as $\theta$ = \{$\theta^{shar}$, $\theta^{pri}$, $\theta^{aux}$\}, where $\theta^{shar}$ corresponds to the shared feature encoder, $\theta^{pri}$ is the primary branch and $\theta^{aux}$ is the auxiliary branch. Note that $\theta^{aux}$ represents the parameters of three auxiliary tasks.

\noindent{\bf Auxiliary Tasks.} Our aim of the auxiliary tasks is to transfer rich and useful knowledge to improve the performance of the primary task. The overall auxiliary loss is the weighted sum of the losses for the three auxiliary tasks:
\begin{equation}
\label{eq:auxloss}
 L_{aux}  =  \lambda_{1} \ell_{rec} + \lambda_{2} \ell_{byol} + \lambda_{3} \ell_{cc}.
\end{equation}

Instead of fixing the values of the balancing weights $\lambda_i$ ($i=1,2,3$), we treat them as learnable parameters and learn their values during training. This allows the learning algorithm to automatically choose the right weights that balance the relative importance of each auxiliary task.
 
To train both primary and auxiliary tasks, we first follow the joint training approach in \cite{ttt}. The loss of the joint training is simply the combination of the primary and auxiliary losses:
\begin{equation}\label{eq:joint}
L_{pri}(\theta^{shar},\theta^{pri}; X, Y, T)+ L_{aux}(\theta^{shar},\theta^{aux}; X, Y).
\end{equation}
Note that since our auxiliary tasks are self-supervised, the auxiliary loss $L_{aux}(\cdot)$ does not need the ground-truth transformation $T$. To simplify the notation, we have assumed one training instance in Eq.~\ref{eq:joint}. It is straightforward to generalize Eq.~\ref{eq:joint} to the entire training set by summing over all training instances. 

The model learned from Eq.~\ref{eq:joint} is then used as the initialization for the meta-auxiliary learning.


\noindent {\bf Primary Task.} We follow the standard learning-based network architecture for point cloud registration. First, a pair of 3D point clouds are passed to a fully convolutional network to extract the corresponding geometric pointwise features. Then the points are matched using the nearest neighbor in the feature space to obtain correspondences. These correspondences are fed to an outlier rejection network which predicts the confidence of each correspondence. Finally, given the correspondences with their associated probability weights resulting from the outlier rejection network, the weighted Procrustes approach is used to align the paired 3D scans by estimating the transformation between the two point clouds. We use $L_{pri}(\theta^{shar},\theta^{pri}; X, Y, T)$ to denote the loss function that measures the difference between the ground-truth transformation $T$ and the prediction $F_{\theta}(X,Y)$.

In this paper, we use Fully Convolutional Geometric Features (FCGF) \cite{fcgf} to extract pointwise features of the 3D point clouds. For the outlier rejection network, we adopt the architecture of three state-of-the-art methods including DGR \cite{dgr}, DHVR \cite{dhvr} and PointDSC \cite{pointDSC}. However, it is importance to note that our proposed meta-auxiliary framework is agnostic to these choices and can be applied to any learning-based point cloud registration methods.


\begin{figure}
    \centering
    \includegraphics[width=0.48\textwidth,height=3.5cm]{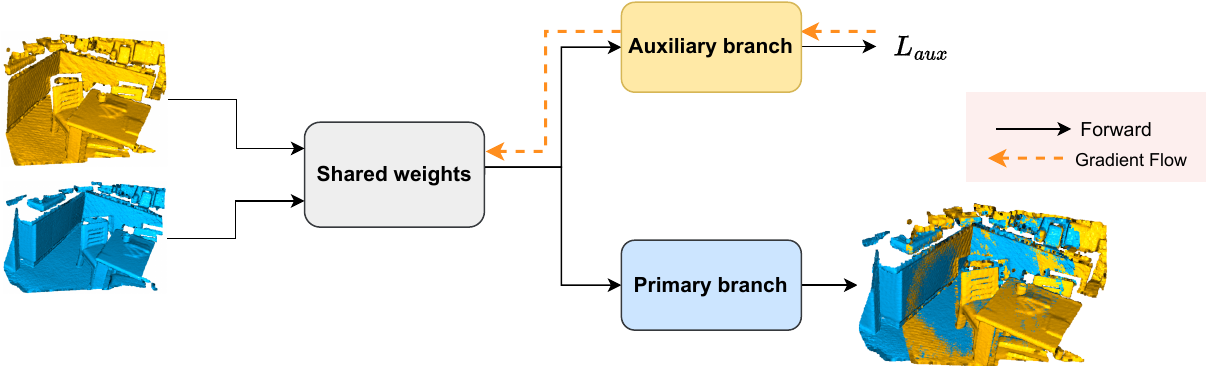}
    \caption{Overview of the proposed meta-auxiliary testing procedure. We use the auxiliary branch to fine-tune the model for each test instance using the auxiliary loss and the adapted model is used to register the input point clouds.}
    \label{fig:Meta-test}
\end{figure}
\subsection{Meta-Auxiliary Learning}

Our goal is to combine self-supervised auxiliary tasks along with the point cloud registration task to quickly adapt the model parameters for each test instance without the need for any extra supervision. Although jointly training the primary and auxiliary tasks will improve the generalization of our model to unknown test distribution, the updated parameters using auxiliary loss may be more biased during training to improve the auxiliary task and not the primary task. Following \cite{deblurring_2021_cvpr}, we propose to use a meta auxiliary task scheme.

\noindent{\bf Training.} We use meta-learning to train the model parameters $\theta$ to be quickly adaptable to different test distribution data, such that updating model parameters at test-time improve the primary point cloud registration task.

Given a batch of paired point clouds $\displaystyle X_{b},\displaystyle Y_{b} $, and the pre-trained model parameters $\theta$ resulting from jointly training primary and auxiliary tasks on the 3DMatch dataset \cite{3Dmatch} by optimizing Eq. \ref{eq:joint}. We perform adaptation for small gradient updates using the auxiliary loss, in which all model parameters ( $\phi^{shar}_{b}, \phi^{pri}_{b},\phi^{aux}_{b}$ ) are updated:

\begin{equation}
\label{eq:innerupdate}
\phi_ {b} \leftarrow \theta - \alpha \nabla_{\theta} L_{aux}(X_{b},Y_{b},\theta),
\end{equation}
where $\alpha$ is the adaptation learning rate. Note that since Eq.~\ref{eq:innerupdate} is based on the auxiliary task, the adaptation can be done at test-time since it does not require the ground-truth transformation.


Then, the adapted model( $\phi^{shar}_{b}, \phi^{pri}_{b}$ ) will be used to perform the primary task and calculate the primary loss. This will enforce the adapted model to boost the primary task performance. The primary loss will be used to optimize the model parameters $\theta$ as:
\begin{equation}
\label{eq:outerupdate}
 \theta \leftarrow \theta - \beta \sum_{b=1}^B \nabla_{\theta} L_{pri}(X_{b},Y_{b},T_{b},\phi_{b}^{shar},\phi_{b}^{pri}),
\end{equation}
where $\beta$ is the meta-learning rate and $B$ is the batch size. Note that $L_{pri}(\cdot)$ in Eq.~\ref{eq:outerupdate} is defined in terms of the updated model $\phi_{b}$ for each instance, while the optimization is performed on the model parameters $\theta$. The training process is summarized in Algorithm \ref{alg:algorithm} and Figure \ref{framework}.

\noindent{\bf Testing.} During test-time, the optimized meta-learned parameters $\theta$ are adapted to a test instance that consists of a pair of 3D point clouds using the auxiliary loss as follows:
\begin{equation}
\phi \leftarrow \theta - \alpha \nabla_{\theta} L_{aux}
\label{eq:tt-update}
\end{equation}
Then, the adapted model ($\phi^{shar}, \phi^{pri}$) is used to perform point cloud registration.

\begin{algorithm}[tb]
\caption{Meta-auxiliary training}
\label{alg:algorithm}
\textbf{Require}: $X,Y,T$: training pairs with their transformation\\
\textbf{Require}: $\alpha,\beta$: learning rates \\
\textbf{Output}: $\theta$: learned parameters
\begin{algorithmic}[1] 
\STATE Initialize the network with pre-trained weights $\theta$
\WHILE{not done}
\STATE Sample a training batch $\{X_{b},Y_{b}, T_{b}\}^B_{b=1}$
\FOR{each example}
\STATE Evaluate the three auxiliary tasks: 
\\ $L_{aux}  =  \lambda_{1} \ell_{rec} + \lambda_{2} \ell_{byol} + \lambda_{3} \ell_{cc}$

\STATE Compute adapted parameters via gradient descent: $\phi_ {b} \leftarrow \theta - \alpha \nabla_{\theta} L_{aux}(X_{b},Y_{b},\theta)$
\STATE Update auxiliary branch: \\ $\theta^{aux} \leftarrow \theta^{aux} - \alpha \nabla_{\theta} L_{aux}(X_{b},Y_{b},\theta^{aux})$
\ENDFOR
\STATE Evaluate the primary task using the adapted parameters and update:
\\ $\theta \leftarrow \theta - \beta \sum_{b=1}^B \nabla_{\theta} L_{pri}(X_{b},Y_{b},T_{b},\phi_{b}^{shar},\phi_{b}^{pri})$
\ENDWHILE
\STATE \textbf{return} $\theta$
\end{algorithmic}
\end{algorithm}

\section{Experiments}

\begin{figure*}
     \centering
     \begin{subfigure}[b]{0.2\textwidth}
         \centering
         \includegraphics[width=\textwidth]{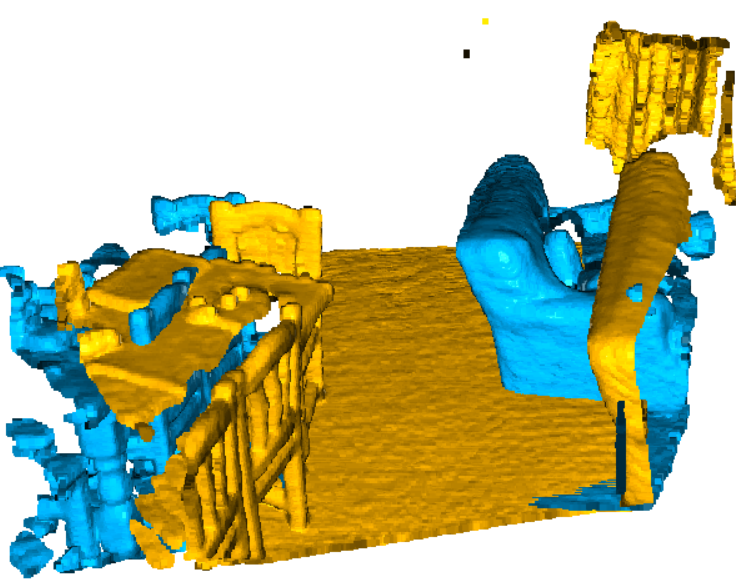}
         \caption*{DGR}
     \end{subfigure}
    \hfill
      \begin{subfigure}[b]{0.19\textwidth}
         \centering
         \includegraphics[width=\textwidth]{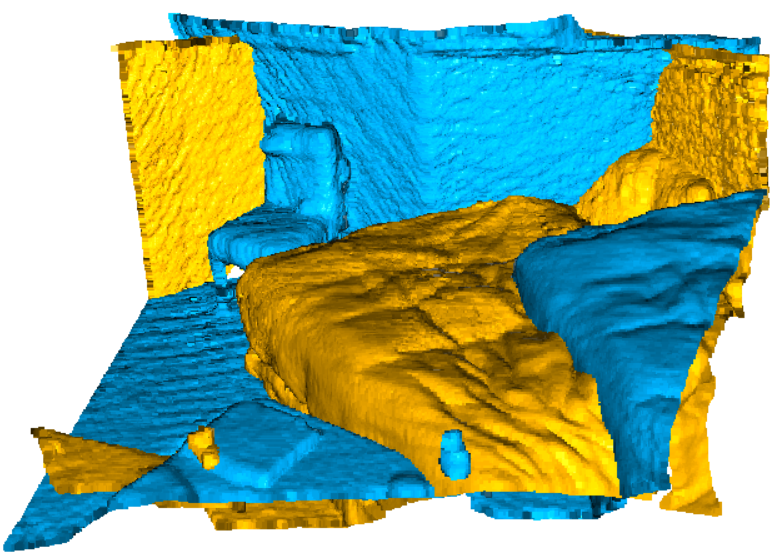}
         \caption*{DHVR }
     \end{subfigure}
     \hfill
     \begin{subfigure}[b]{0.19\textwidth}
         \centering
         \includegraphics[width=\textwidth]{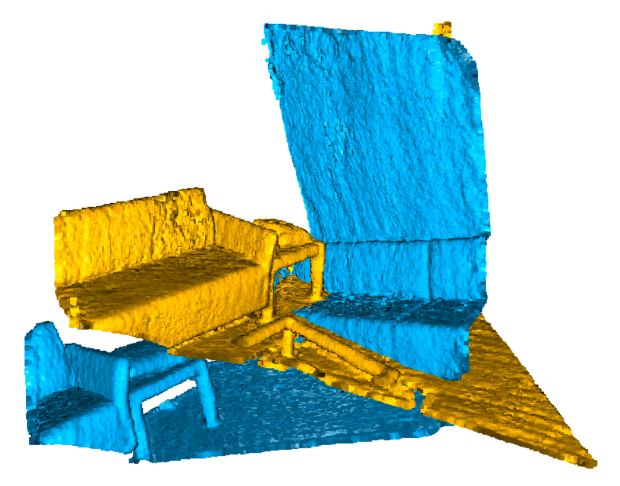}
         \caption*{PointDSC}
     \end{subfigure}
     \hfill
     \begin{subfigure}[b]{0.2\textwidth}
         \centering
         \includegraphics[width=\textwidth]{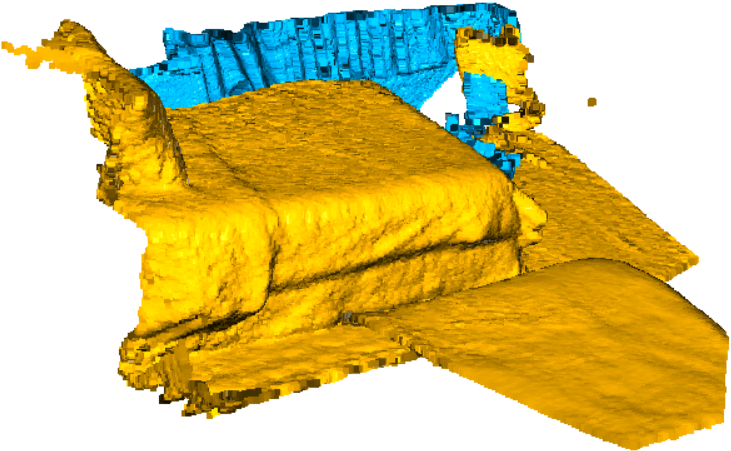}
         \caption*{DGR}
     \end{subfigure}
     \hfill
     \begin{subfigure}[b]{0.19\textwidth}
         \centering
         \includegraphics[width=\textwidth]{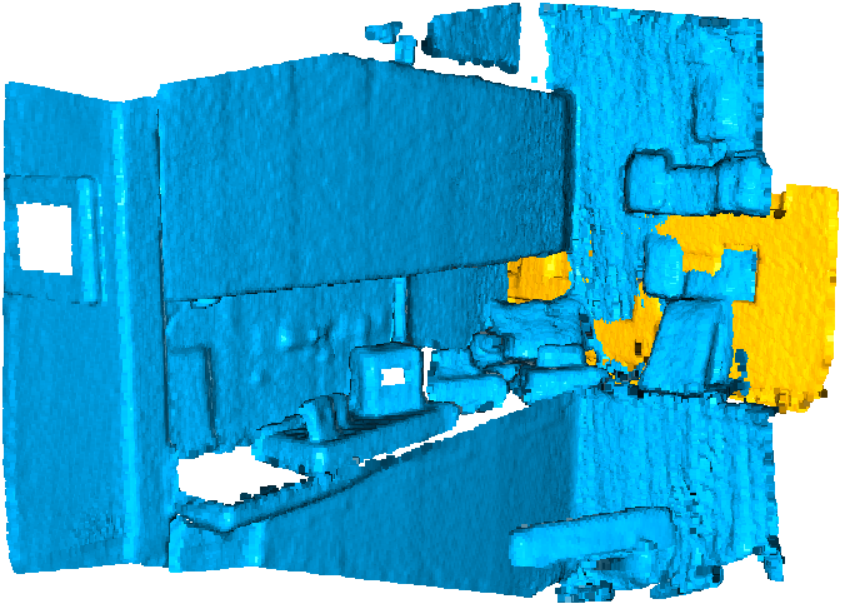}
         \caption*{PointDSC}
     \end{subfigure}
     \hfill
     \begin{subfigure}[b]{0.19\textwidth}
         \centering
         \includegraphics[width=\textwidth]{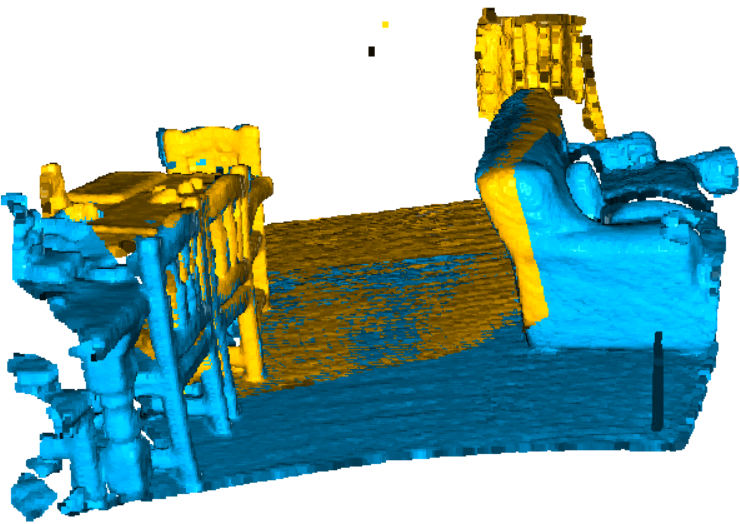}
         \caption*{Ours + DGR}
     \end{subfigure}
      \hfill
     \begin{subfigure}[b]{0.2\textwidth}
         \centering
         \includegraphics[width=\textwidth]{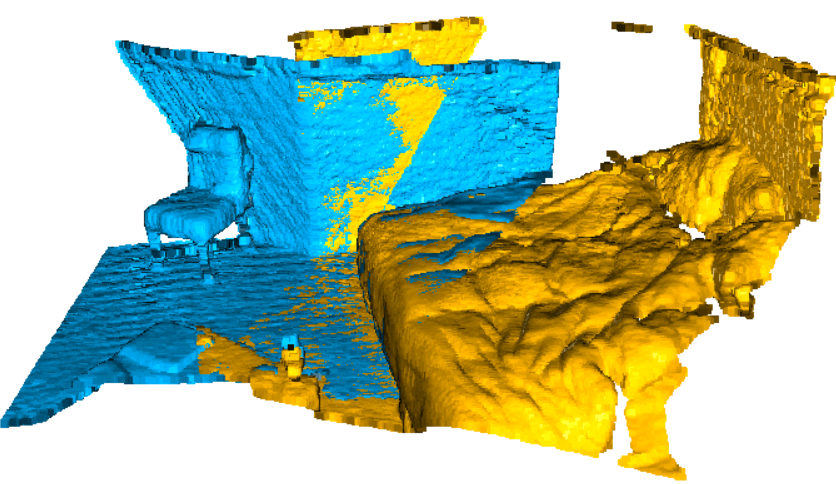}
         \caption*{Ours + DHVR}
     \end{subfigure}
     \hfill
     \begin{subfigure}[b]{0.19\textwidth}
         \centering
         \includegraphics[width=\textwidth]{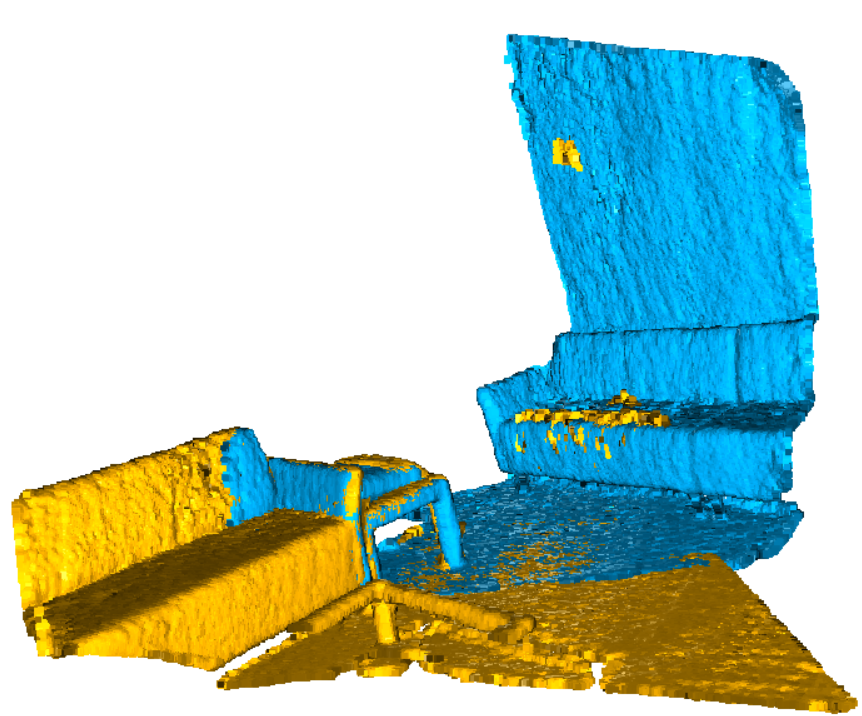}
         \caption*{Ours + PointDSC}
     \end{subfigure}
     \hfill
      \begin{subfigure}[b]{0.19\textwidth}
         \centering
         \includegraphics[width=\textwidth]{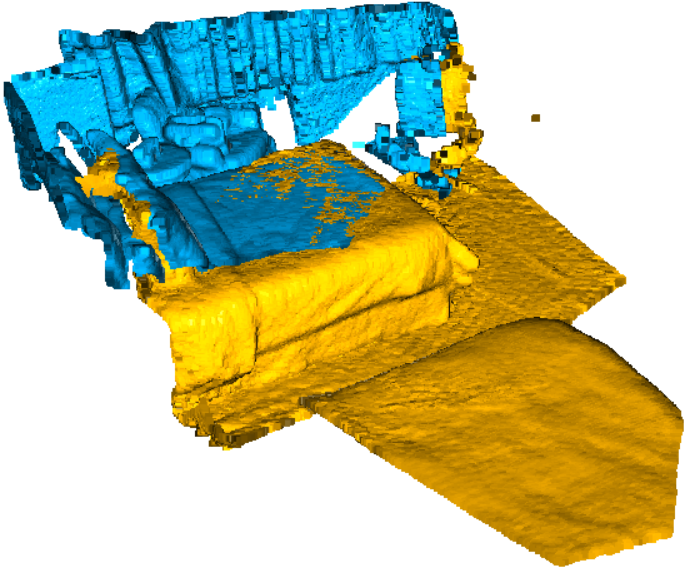}
         \caption*{Ours + DGR}
     \end{subfigure}
     \hfill
     \begin{subfigure}[b]{0.21\textwidth}
         \centering
         \includegraphics[width=\textwidth]{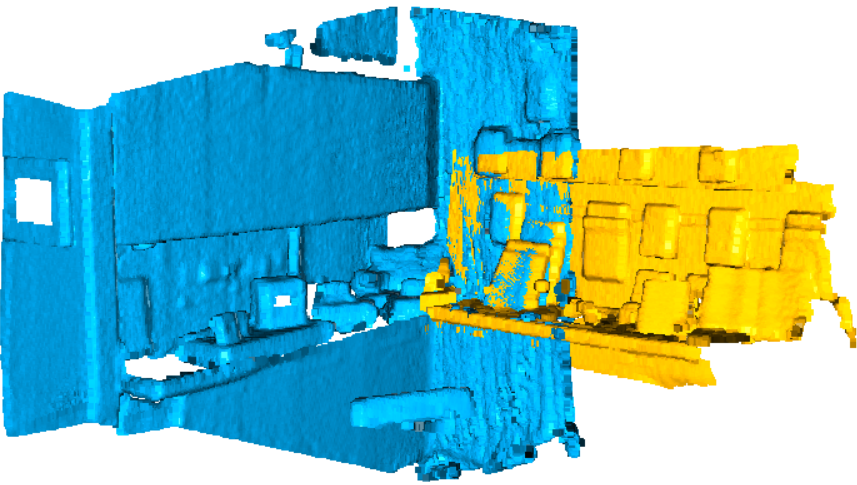}
         \caption*{Ours + PointDSC}
     \end{subfigure}
     \hfill
        \caption{We propose a generic test-time adaptation framework that can be applied to boost standard point cloud registration pipeline. Here we show qualitative comparisons between the baselines and our method. The first three columns are from 3DMatch dataset \cite{3Dmatch} and the last two columns are from 3DLoMatch dataset \cite{geoTrans}. Our proposed framework can successfully align failure examples of DGR \cite{dgr}, DHVR \cite{dhvr}, and PointDSC \cite{pointDSC}.}
\label{fig:qualitative}
\end{figure*}
We first evaluate our method on a 3D indoor dataset for the pairwise registration task. Then we analyze the generalization of our proposed model to unseen 3D outdoor datasets. Additionally, we integrate our method into a multi-way registration pipeline and evaluate its performance on generating final 3D reconstruction scenes. Finally, we perform extensive ablation studies to inspect each component of our approach. Additional results and ablation studies are provided in the supplementary document.

\subsection{Experimental Setup}
\paragraph{Dataset.} We use the 3DMatch benchmark \cite{3Dmatch} for indoor pairwise registration. It consists of point cloud pairs with corresponding ground-truth transformations from real-world indoor scenes scanned by commodity RGB-D sensors. We follow the standard splitting strategy and evaluation protocol in 3DMatch \cite{3Dmatch}, where the test data contain 1623 partially overlapping 3D point cloud scans from 8 different indoor scenes. For the outdoor dataset, we use the KITTI odometry benchmark \cite{kitti} which consists of 3D outdoor scenes scanned using a Velodyne laser. We follow the train/test split in \cite{fcgf} to create pairwise splits, since the official benchmark does not have labels for pairwise registration. We perform voxel downsampling to generate point clouds with uniform density and set voxel size to 5cm for indoor dataset and 30cm for outdoor dataset.  For multi-way registration experiment, we use the simulated Augmented ICL-NUIM dataset \cite{icl} which contains augmented indoor reconstruction scenes from RGB-D videos.

\paragraph{Evaluation Metrics.}
Following \cite{dgr,pointDSC}, we report Registration Recall (RR), Rotation Error (RE) and Translation Error (TE). RE and TE are defined as: 
\begin{equation}
RE = \arccos\frac{Tr(R^TR^*)-1}{2}, TE = ||t-t^*||^2,
\end{equation}
where $R^*$ and $t^*$ are the ground-truth rotation and translation, respectively. Registration Recall (RR) is the ratio of successful pairwise registration that its rotation error and translation error are below predefined thresholds. These thresholds are set to ($RE=15$, $TE =30cm$) for indoor scenes and ($RE=5$, $TE =60cm$) for outdoor scenes.

\paragraph{Implementation Details.}
We implement our framework in PyTorch and use the official implementation of DGR \cite{dgr}, DHVR \cite{dhvr}, and PointDSC \cite{pointDSC} as the backbones of our approach. We first jointly train primary and auxiliary tasks by optimizing the loss in Eq. \ref{eq:joint} using the ADAM optimizer with an initial learning rate of $10^{-4}$ and an exponentially decayed factor of $0.99$. For meta-training, the learning rates $\alpha$ and $\beta$ are set to $2.5\times 10e^{-5}$. We perform 5 gradient updates during training and testing to adapt the model parameters using the auxiliary loss in Eq. \ref{eq:auxloss}. All experiments are conducted on an NVIDIA TitanX GPU.
\begin{table}
\caption{Comparison with other state-of-the-art methods on the 3DMatch dataset \cite{3Dmatch}. $\uparrow$ ( or $\downarrow$) indicates that a higher (or lower) number means better performance.}
\label{3Dmatch-results}
\centering
\begin{small}
\begin{tabular}{l|lll}
\hline
  & Recall $\uparrow$ & RE (deg) $\downarrow$ & TE (cm) $\downarrow$ 
 \\
\hline\hline
FGR \cite{fgr} & 78.56 & 2.82 & 8.36 \\
TEASER \cite{teaser} & 85.77 & 2.73 & 8.66 \\
GC-RANSAC \cite{GCransac} & 92.05 & 2.33 & 7.11 \\
RANSAC-1M \cite{ransac} & 88.42 & 3.05 & 9.42 \\
RANSAC-2M \cite{ransac} & 90.88 & 2.71 & 8.31 \\
RANSAC-4M  \cite{ransac} & 91.44 & 2.69 & 8.38 \\
CG-SAC \cite{cgsac}& 87.52 & 2.42 & 7.66 \\
LEAD \cite{lead} & 67.15 & 3.39 &  12.01 \\
Ppf-foldnet \cite{ppf} & 71.82  & 3.45 & 9.16  \\
3DRegNet \cite{3dreg} & 77.76 & 2.74 & 8.13  \\
\hline\hline
DGR \cite{dgr} & 91.30 & 2.40 & 7.48 \\
\textbf{Ours + DGR} & \textbf{92.45} & \textbf{1.71} & \textbf{6.39} \\
\hline
DHVR \cite{dhvr} & 91.40 & 2.08 & 6.61 \\
\textbf{Ours + DHVR} & \textbf{92.28} & \textbf{1.75} & \textbf{6.42} \\
\hline
PointDSC \cite{pointDSC} & 92.85 & 2.08 & 6.51  \\
PointDSC-reported  & 93.28 & 2.06 & 6.55  \\
\textbf{Ours + PointDSC} & \textbf{93.47} & \textbf{1.70} & \textbf{6.21} \\
\hline
\end{tabular}
\end{small}
\end{table}
\begin{figure*}
     \centering
     \begin{subfigure}[b]{0.485\textwidth}
         \centering
         \includegraphics[width=\textwidth]{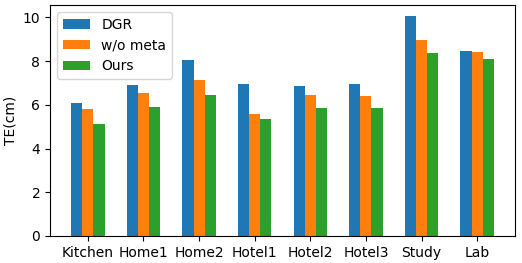}
     \end{subfigure}
     \hfill
     \begin{subfigure}[b]{0.485\textwidth}
         \centering
         \includegraphics[width=\textwidth]{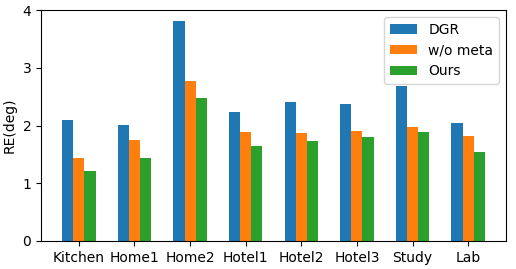}
     \end{subfigure}
        \caption{Registration results for different scenes on the 3DMatch dataset in terms of RE and TE. ``w/o meta" refers to a variant of our method that simply optimizes Eq. \ref{eq:joint} during training without adopting the meta-auxiliary training paradigm.
}
\label{fig:scenes_bar}
\end{figure*}
\subsection{Main Results}
We first evaluate our method on the 3DMatch dataset and report the results in Table \ref{3Dmatch-results}. We compare our method with 5 traditional methods: FGR \cite{fgr}, TEASER \cite{teaser}, GC-RANSAC \cite{GCransac}, RANSAC \cite{ransac}, CG-SAC \cite{cgsac}, 2 unsupervised learning-based methods: LEAD \cite{lead}, Ppf-foldnet \cite{ppf}, and 4 supervised learning-based methods: 3DRegNet \cite{3dreg}, DGR \cite{dgr}, DHVR \cite{dhvr}, PointDSC \cite{pointDSC}. All learning-based methods are trained on the 3DMatch dataset and follow the same experiment setup for fair comparison. As shown in Table \ref{3Dmatch-results}, our method improves the registration recall of DGR \cite{dgr} and DHVR \cite{dhvr} by about 1\%, and PointDSC \cite{pointDSC} by about 0.5\%, as well as the RE and TE have significantly decreased for all three backbones (on average 7.3\% and 21\%). More importantly, our method with PointDSC \cite{pointDSC} as backbone outperforms all other state-of-the-art methods. Figure \ref{fig:qualitative} shows qualitative results comparison on challenging examples of 3DMatch \cite{3Dmatch} and 3DLoMatch \cite{geoTrans} datasets when applying our TTA method to DGR \cite{dgr}. Our meta-auxiliary framework enables the model to better capture the internal features of each test-instance, leading to performance improvement.

\begin{figure}[t]
\centering
\includegraphics[width=\linewidth,height=3.2cm]{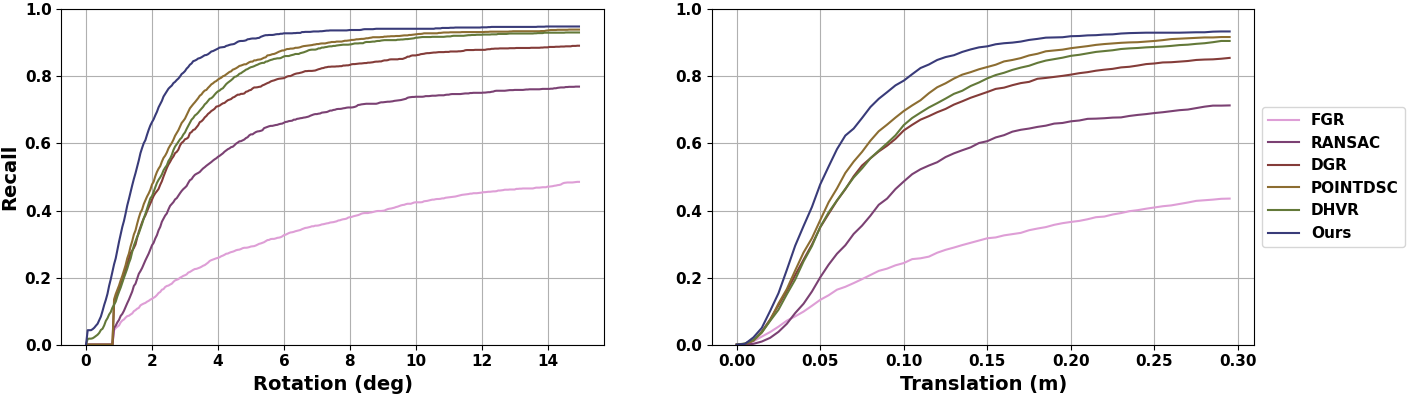}
   \caption{Comparison of the registration recall on the 3DMatch dataset between our approach and other state-of-the-art methods by varying the translation error and the rotation error thresholds. Our framework with PointDSC as backbone outperforms all other methods for all thresholds.}
\label{fig:recall TE/RE}
\end{figure}

Figure \ref{fig:scenes_bar} shows the 3DMatch registration results per scene. As an ablation study, we remove the meta learning diagram and simply optimize the joint loss Eq. \ref{eq:joint} during training\footnote{At test-time, we still perform Eq. \ref{eq:tt-update} for TTA.}. By doing so, we obtain superior results than the backbone DGR \cite{dgr}, which demonstrates that our proposed auxiliary tasks can complement the primary PCR task and thus improves accuracy. Moreover, our final meta-auxiliary framework achieves the best. Figure \ref{fig:recall TE/RE} shows the robustness of our approach under different rotation and translation error thresholds.

\begin{table}[t]
\caption{Results of cross-dataset generalization. Here we train the model on 3DMatch \cite{3Dmatch} and evaluate on KITTI \cite{kitti}. The results of training on KITTI and evaluating on 3DMatch can be found in the supplementary document. }
\label{generalization}
\centering
\begin{small}
\begin{tabular}{l|lll}
\hline
& Recall $\uparrow$ & RE (deg) $\downarrow$ & TE (cm) $\downarrow$  \\
\hline
DGR  & 95.24 & 0.44 & 23.25 \\
\textbf{Ours + DGR} & \textbf{97.36} & \textbf{0.34} & \textbf{21.16} \\
\hline
DHVR & 95.82 & 0.39 &  22.17  \\
\textbf{Ours + DHVR} & \textbf{98.01}& \textbf{0.32} & \textbf{21.18} \\
\hline
PointDSC  & 97.15 & 0.36 & 21.74 \\
\textbf{Ours + PointDSC} & \textbf{98.23} & \textbf{0.33} & \textbf{20.86}\\
\hline
\end{tabular}
\end{small}
\end{table}

\subsection{Ablation and Additional Results}
We perform additional experiments and ablation studies to further analyze our proposed method.

\noindent{\bf Outdoor Registration Generalization.} 
Although our method achieves the best performance over all the traditional and learning-based methods, the main advantage of our method is the ability to generalize to unseen test distribution. In order to evaluate the generalization of our method, we perform a cross-dataset experiment on both 3DMatch \cite{3Dmatch} and KITTI \cite{kitti} datasets, where the trained model on 3DMatch \cite{3Dmatch} is used to test on KITTI \cite{kitti} and vice versa. As shown in Table \ref{generalization}, our method shows a significant improvement on all evaluation metrics when training on 3DMatch dataset \cite{3Dmatch} and evaluating on KITTI dataset \cite{kitti}, demonstrating the effectiveness of the proposed framework. To further inspect the generalization of our method, we test our framework on another widely-used outdoor dataset, namely ETH dataset \cite{ETH_data}. In Table \ref{ETH-results}, we report the registration results of our approach when evaluating on ETH dataset \cite{ETH_data}. We can observe that our method improves the performance of the baselines across all scenes with a good margin. In particular, the average recall of DGR \cite{dgr} is significantly improved by about 5\%, which sheds light on the generalization capability of our approach.

\begin{table}
\caption{Registration results on ETH dataset \cite{ETH_data}. We report the registration recall per scene as well as the average recall across all scenes.}
\label{ETH-results}
\centering
\begin{small}
\resizebox{\linewidth}{!}{%
\begin{tabular}{l|cccc|ccc}
\hline
& \multicolumn{2}{c}{ \textbf{Gazebo}} & \multicolumn{2}{c}{ \textbf{Wood}} & \multirow{2}{*}{\textbf{AVG}}\\
  & Summer & Winter &  Summer & Autumn\\
\hline
DGR  & 92.63 & 83.28 & 79.86 &  71.34 & 81.78 \\
\textbf{Ours + DGR} & \textbf{95.28} & \textbf{87.95} & \textbf{81.73} & \textbf{82.45} & \textbf{86.85} \\
\hline
DHVR  & 93.78 &  82.74 & 81.07  & 75.32  & 83.22  \\
\textbf{Ours + DHVR} & \textbf{95.96} & \textbf{87.13} & \textbf{82.25} & \textbf{80.79} & \textbf{86.53} \\
\hline
PointDSC & 94.21 & 89.68  &  83.14 & 78.42 & 86.36 \\
\textbf{Ours + PointDSC}  & \textbf{94.89} & \textbf{91.49} & \textbf{87.02} & \textbf{85.22} & \textbf{89.65} \\  
\hline
\end{tabular}%
}
\end{small}
\end{table}

\begin{figure}
\begin{center}
\includegraphics[width=0.3\textwidth,height=4.5cm]{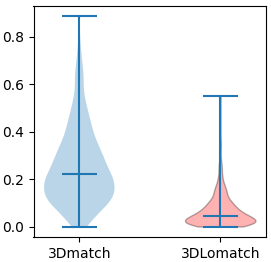}
\end{center}

\caption{Comparison between the distribution of the inlier ratio of correspondences obtained by feature matching on 3DMatch \cite{3Dmatch} and 3DLoMatch \cite{geoTrans} benchmarks. The registration task is more challenging with a lower inlier ratio.}
\label{fig:data distribution}
\end{figure}

\noindent{\bf Robustness to Low-Overlapping Point Clouds.} To further validate the robustness of our method, we evaluate our method on a dataset with low-overlapping ratio between input point clouds, namely 3DLoMatch \cite{geoTrans}. This dataset is constructed from the 3DMatch benchmark \cite{3Dmatch} and has a low-overlapping ratio (10\%-30\%) between 3D point cloud fragments. Figure \ref{fig:data distribution} compares the inlier ratio between 3DLoMatch \cite{geoTrans} and 3DMatch \cite{3Dmatch}, which shows that 3DLoMatch \cite{geoTrans} is more challenging due to the lower inlier ratio. We use the model trained on 3DMatch \cite{3Dmatch} for evaluation and report our results in Table \ref{lowoverlap}. Our approach outperforms all other methods, demonstrating the robustness of our approach to low-overlapping scenarios. More importantly, this validates the robustness of our approach to the percentage of template and target overlaps where 3Dmatch \cite{3Dmatch} contains overlapping ratios ($\geq$30\%) and 3DLoMatch \cite{geoTrans} contains low-overlapping ratios (10\%-30\%). The evaluation results on both datasets show the superiority of our approach among the baselines under different ratios.

\begin{table}
\caption{Robustness to low-overlapping point clouds on the 3DLoMatch dataset \cite{geoTrans} with low-overlapping ratio between 3D point cloud segments. We train on 3DMatch and evaluate on 3DLoMatch.}
\label{lowoverlap}
\centering
\begin{small}

\begin{tabular}{l|lll}
\hline
& Recall $\uparrow$ & RE (deg) $\downarrow$ & TE (cm) $\downarrow$  \\
\hline
DGR  & 43.80 & 4.17 & 10.82 \\
\textbf{Ours + DGR} & \textbf{50.73} & \textbf{4.06}  & \textbf{10.52}\\
\hline
DHVR  & 54.46  & 4.13 & 10.54  \\
\textbf{Ours + DHVR} & \textbf{57.32}  & \textbf{3.83} & \textbf{10.26}\\
\hline
PointDSC  & 56.10  & 3.87 & 10.39\\
\textbf{Ours + PointDSC} & \textbf{57.81}  & \textbf{3.79} & \textbf{10.15} \\
\hline
\end{tabular}
\end{small}
\end{table}

\begin{table}
\caption{Multiway registeration results on Augmented ICL-NUIM dataset evaluated by ATE(cm) where lower is better.}
\label{Multi-way}
\centering
\begin{small}
\resizebox{\linewidth}{!}{%
\begin{tabular}{l|lllll}
\hline
  & Living1 & Living2 & Office1 & Office2 & AVG \\
\hline
FGR  & 78.97 & 24.91 & 14.96 & 21.05 & 34.98 \\
RANSAC  & 110.9 & 19.33 & 14.42 & 17.31 & 40.49 \\
\hline
DGR  & 21.06 & 21.88 & 15.76 & 11.56 & 17.57 \\
\textbf{Ours + DGR} & \textbf{18.32}& \textbf{16.12} & \textbf{12.24} & \textbf{10.44} & \textbf{14.28} \\
\hline
DHVR  & 22.91 & 16.37 & 12.58 & 10.90 & 15.69 \\
\textbf{Ours + DHVR} & \textbf{18.46} & \textbf{13.59}  & \textbf{12.43} & \textbf{9.56} & \textbf{13.51}\\
\hline
PointDSC  & 20.25 & 15.58 & 13.56 & 11.30 & 15.18 \\
\textbf{Ours + PointDSC}    & \textbf{15.73} & \textbf{12.07} &  \textbf{12.15} & \textbf{9.78} & \textbf{12.43}\\  
\hline
\end{tabular}%
}
\end{small}
\end{table}

\begin{table}
\caption{Ablation studies on framework components: Auxiliary Learning, Meta Learning, and Test-time Adaptation.}

\label{ablation_components}
\centering
\begin{small}
\resizebox{\linewidth}{!}{%
\begin{tabular}{l|lll}
\hline
& Recall $\uparrow$ & RE (deg) $\downarrow$ & TE (cm) $\downarrow$ \\
\hline
DGR  & 91.31  & 2.40 & 7.48  \\
DGR + Aux. & 91.42  & 2.25  & 7.06 \\
DGR  + TTA (w/o meta) & 91.86  & 1.88 & 6.54   \\
DGR + Meta-Aux. (w/o TTA)  &  92.28 &  1.71 &6.40   \\
\textbf{DGR + full framework}  & \textbf{92.45} &  \textbf{1.71} &\textbf{6.39} \\
\hline
\end{tabular}%
}
\end{small}
\end{table}

\begin{table}
\caption{Ablation studies on the three auxiliary tasks: Point Cloud Reconstruction (rec), Correspondence Classification (cc), and Feature Learning (byol).}

\label{ablation_tasks}
\begin{center}
\begin{small}
\begin{tabular}{l|lll}
\hline
& Recall $\uparrow$  & RE (deg) $\downarrow$ & TE (cm) $\downarrow$ \\
\hline
DGR \cite{dgr} & 91.31 & 2.43 & 7.34  \\

DGR + rec & 92.24 & 1.71  & 6.42  \\

DGR  + (rec, cc) & 92.38 & \textbf{1.69} & 6.40   \\

\textbf{DGR + (rec, cc, byol)}  & \textbf{92.45} & 1.71 & \textbf{6.39} \\

\hline
\end{tabular}
\end{small}
\end{center}

\end{table}

\noindent{\bf Multiway Registration for 3D Reconstruction.} Point cloud registration is a critical step for various 3D applications. In this section, we present the effect of pairwise registration performance on obtaining more accurate and robust 3D reconstruction scenes. Following \cite{dgr,pointDSC}, we integrate our method into a 3D reconstruction pipeline\cite{icl}. Given RGB-D scans, 3D fragments are generated from the scene. Next, we perform pairwise registration using our method to align all fragments. Finally, multi-way registration \cite{icl} is used to optimize the fragment poses using pose graph optimization \cite{poseGraph}. We use the model trained on 3DMatch to further demonstrate the generalization of our method and evaluate our approach on Augmented ICL dataset using Absolute Trajectory Error (ATE). As shown in Table \ref{Multi-way}, our method achieves the lowest error compared to all other methods.


\noindent{\bf{Methodology Components.}}
To study the effectiveness of the proposed framework, we conduct ablation experiments on 3DMatch dataset \cite{3Dmatch} and evaluate the effect of each component of the proposed framework. We consider DGR \cite{dgr} as the backbone in our experiments and report the results after applying each component to DGR \cite{dgr}. Specifically, we compare the results between our method's three major components: Auxiliary Learning, Meta Learning, Test-time Adaptation. We first investigate the effect of our proposed Auxiliary Learning method by jointly training the primary and three auxiliary tasks by optimizing the loss in Eq. \ref{eq:joint}. This shows the strength of the proposed auxiliary tasks acting as a regularizer during training. Then, we study the impact of combining test-time adaptation with auxiliary learning, in which the auxiliary tasks are used to update the model parameters at test-time by optimizing the auxiliary loss in Eq. \ref{eq:auxloss}. Furthermore, we show the effect of the proposed meta-auxiliary learning paradigm elaborated in Algorithm \ref{alg:algorithm} in learning optimal model parameters. However, we fixed the model parameters at test-time. Finally, we report the results of our final framework integrating Auxiliary Learning, Meta Learning, and Test-time Adaptation.

As reported in Table \ref{ablation_components}, auxiliary learning improves the registration results across all evaluation metrics compared to DGR \cite{dgr}. This demonstrates that the proposed auxiliary tasks can complement the registration task, leading to performance improvement. Combining auxiliary learning with TTA has further improved the performance of registration recall by 0.44\%, and decreased the TE and RE by 0.52cm and 0.37 deg, respectively. As TTA allows the auxiliary tasks to transfer useful features of test-instance to the primary registration task, it enhances the registration performance. Moreover, the proposed meta-auxiliary training method greatly boosts the performance, which demonstrates the effectiveness of training tasks using meta-learning terminology such that the meta-objective enforces the auxiliary tasks to improve the primary task performance. Finally, our final framework further boosts the registration performance by fine-tuning the model parameters at test-time.

\noindent{\bf{Analysis of Auxiliary Tasks.}}
We conduct an additional ablation study on the 3DMatch dataset \cite{3Dmatch} to investigate the importance of each auxiliary task in our approach in improving the registration performance. As shown in Table \ref{ablation_tasks}, the auxiliary reconstruction task significantly boosts the registration recall of DGR \cite{dgr} by 0.92\%. Also, Translation Error (TE) and Rotation Error (RE) greatly drop by 13\% and 30\%, respectively. These evaluation metrics are further improved when combining the auxiliary correspondence classification task to the reconstruction task. This demonstrates the impact of multiple auxiliary tasks in transferring additional features to the primary task and enhancing registration results. Finally, our final three auxiliary tasks achieve a higher registration recall of 92.45\% and a lower Translation Error (TE) of 6.39cm. However, the Rotation Error (RE) was slightly worse when compared to the two auxiliary tasks results.

\section{Conclusion}
We have introduced a novel test-time adaptation framework for point cloud registration using multitask meta-auxiliary learning. Previous work usually follows a supervised learning approach to train a model on a labeled dataset and fix the model during evaluation on unseen test data. In contrast, our framework is designed to effectively adapt the model parameters at test time for each test instance to boost the performance. We have introduced three self-supervised auxiliary tasks to improve the the primary registration task. Furthermore, we have used a meta-auxiliary learning paradigm to train the primary and auxiliary tasks, so that the adapted model using auxiliary tasks improve the performance of the primary task. Extensive experiments show the effectiveness of the proposed approach in improving the registration performance and outperforming state-of-the-art methods.

{\small
\bibliographystyle{ieee_fullname}
\bibliography{egbib}
}
\section{Supplementary Material}
\subsection{Learnable Balancing Weights}
The final auxiliary loss is the weighted sum of our proposed three auxiliary losses. Instead of using the same fixed values of the auxiliary losses balancing weights, we propose to add the balancing weights to the learnable network parameters and learn their values during training. This enables the training algorithm to effectively balance the auxiliary tasks with optimal weights. At test-time, the learned weights are fixed.

We first initialize three trainable parameters $c_{i}(i=1,2,3)$ with initial values of ones. These parameters are trained along with the auxiliary task parameters, i.e. $\{\theta^{aux}, c_i\}$, by optimizing the final auxiliary loss $L_{aux}$. To ensure that the learned balancing weights are in the appropriate range, we use the loss function in Eq. \ref{eq:learn} to avoid large values of weights and biased losses. We further apply a softmax function to make balancing weights sum to 1. The learned parameters $c_{i}$ are mapped to $\lambda_{i}$ as following:

\begin{equation}
\label{eq:learn}
 \lambda_{i} = Softmax (\frac{1}{2c_{i}^2}), 
\end{equation}
where i = (1,2,3). The final balanced auxiliary loss is defined as following:
\begin{equation}
\label{eq:aux}
 L_{aux}  (\{c_i\}_{i=1}^3, \theta_{aux})=  \lambda_{1} \ell_{rec} + \lambda_{2} \ell_{byol} + \lambda_{3} \ell_{cc}.
\end{equation}

\subsection{Additional Results and Ablation Studies}

\noindent{\bf Generalization to Unseen Datasets.} 
We have conducted a cross-dataset generalization experiment on 3DMatch \cite{3Dmatch} and KITTI \cite{kitti} datasets to evaluate the capability of our method in improving the generalization performance of point cloud registration networks. Due to the space limitation, we only report the registration results when training the model on 3DMatch \cite{3Dmatch} and evaluating on KITTI \cite{kitti} in the main paper. Here we report the performance improvement of all the baselines when training on KITTI dataset \cite{kitti} and testing on 3DMatch dataset \cite{3Dmatch}. As presented in Table \ref{generalization_results}, our method achieves the best performance across all the evaluation metrics with a good margin. This demonstrates the effectiveness of our method in boosting the generalization capability of the three backbones: DGR \cite{dgr}, DHVR \cite{dhvr}, and PointDSC \cite{pointDSC} to unseen datasets by enabling the networks to exploit the internal features of point clouds at test time.

\begin{table}
\caption{Results of cross-dataset generalization experiment. We train the models on KITTI dataset \cite{kitti} and evaluate on 3DMatch dataset  \cite{3Dmatch}.} 
\label{generalization_results}
\centering
\begin{small}
\begin{tabular}{l|lll}
\hline
& Recall $\uparrow$ & RE (deg) $\downarrow$ & TE (cm) $\downarrow$  \\
\hline
DGR \cite{dgr} & 87.39& 2.71 &7.58 \\
\textbf{Ours + DGR} & \textbf{90.25} & \textbf{2.32} & \textbf{7.26} \\
\hline
DHVR \cite{dhvr} & 87.16 & 2.74 & 7.43  \\
\textbf{Ours + DHVR} & \textbf{90.48}& \textbf{2.25} & \textbf{7.04} \\
\hline
PointDSC \cite{pointDSC} &89.42 & 2.15 & 6.89 \\
\textbf{Ours + PointDSC} & \textbf{91.36} & \textbf{1.87} & \textbf{6.33}\\
\hline
\end{tabular}
\end{small}
\end{table}
\begin{table}
\caption{Ablation studies comparison between learnable and fixed balancing weights. Second row presents results with fixed balancing weights of ($\lambda_1=0.5$, $\lambda_2=0.3$, $\lambda_3=0.2$). Third row presents results with fixed balancing weights of ($\lambda_1=0.7$, $\lambda_2=0.1$, $\lambda_3=0.2$). Our framework with learnable balancing weights achieve better results in all evaluation metrics.}

\vspace{-5mm}
\label{ablation_learnable}
\begin{center}
\begin{small}
\begin{tabular}{l|lll}
\hline
& Recall $\uparrow$  & RE (deg) $\downarrow$ & TE (cm) $\downarrow$ \\
\hline
DGR \cite{dgr}  & 91.31  & 2.43 & 7.34  \\

DGR  + fixed & 92.03 & 1.76 & 6.48  \\

DGR + fixed & 92.26 & 1.73  & 6.41 \\

\textbf{DGR + learnable}  & \textbf{92.45} & \textbf{1.71} & \textbf{6.39}\\

\hline
\end{tabular}
\end{small}
\end{center}
\end{table}

\noindent{\bf Analysis of methodology components.} 
We perform ablation study on each methodology component to further analyze our proposed method. We report the registration results when evaluating on 3DMatch dataset \cite{3Dmatch} in the main paper. Here we report results when testing on KITTI dataset \cite{kitti} to investigate the contribution of each framework component. As shown in Table \ref{ablation_comp}, the results of combining auxiliary learning with DGR \cite{dgr} were slightly worse than baseline results. However, TTA and meta-learning significantly improve registration performance when combined with auxiliary learning. Finally, our final framework achieves the best registration results, which validate the contribution of each component.
\begin{table}
\caption{Ablation study on KITTI dataset. We report the contribution of each framework component.}
\label{ablation_comp}
\centering
\begin{small}
\resizebox{\linewidth}{!}{%
\begin{tabular}{l|lll}
\hline
& Recall $\uparrow$ & RE (deg) $\downarrow$ & TE (cm) $\downarrow$ \\
\hline
DGR  & 95.24 & 0.44 & 23.25 \\
DGR + Aux. & 95.21  & 0.43  & 23.32\\
DGR  + TTA (w/o meta) & 96.63 & 0.39 & 22.30   \\
DGR + Meta-Aux. (w/o TTA)  & 96.85 & 0.37 & 21.94 \\
\textbf{DGR + full framework}  & \textbf{97.36}  & \textbf{0.34} & \textbf{21.16} \\
\hline
\end{tabular}%
}
\end{small}
\vspace{-5mm}
\end{table}

\noindent{\bf Impact of Learnable Balancing Weights.} 
In this study, we report the impact of learning the balancing weights. We perform the experiments on the 3DMatch dataset and adapt DGR \cite{dgr} as the baseline of the experiment. The results are shown in Table \ref{ablation_learnable}. In the second and third rows of Table \ref{ablation_learnable}, we report the results of using fixed balancing weights. Although the registration recall improved by 0.72\% and 0.95\%, respectively, it is hard to determine the optimal balancing weights without doing numerous experiments. Instead, training with learnable balancing weights effectively balances the auxiliary tasks and greatly improves all evaluation metrics.

\subsection{More Qualitative Results}
\begin{figure*}
\centering
      \begin{subfigure}[b]{0.26\textwidth}
         \centering
        \includegraphics[width=\textwidth]{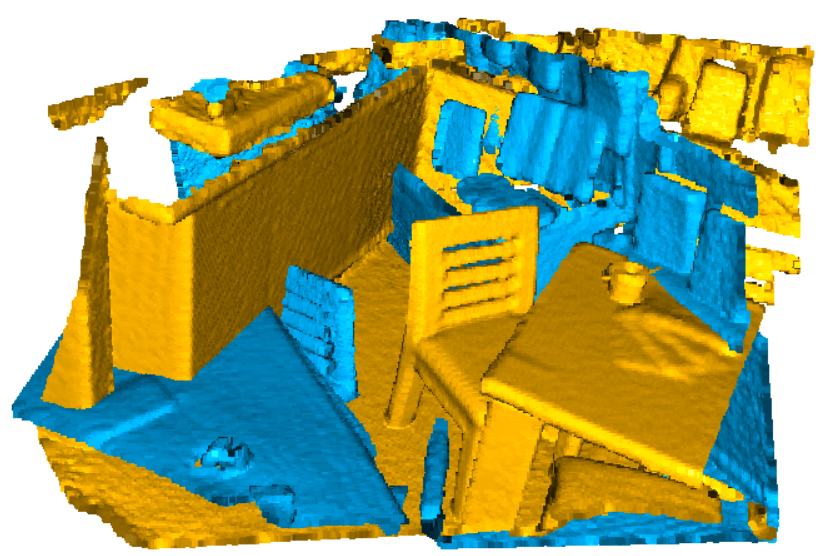}
    \caption*{DGR}
     \end{subfigure}
     \hspace{0.5em}
      \begin{subfigure}[b]{0.26\textwidth}
         \centering
         \includegraphics[width=\textwidth]{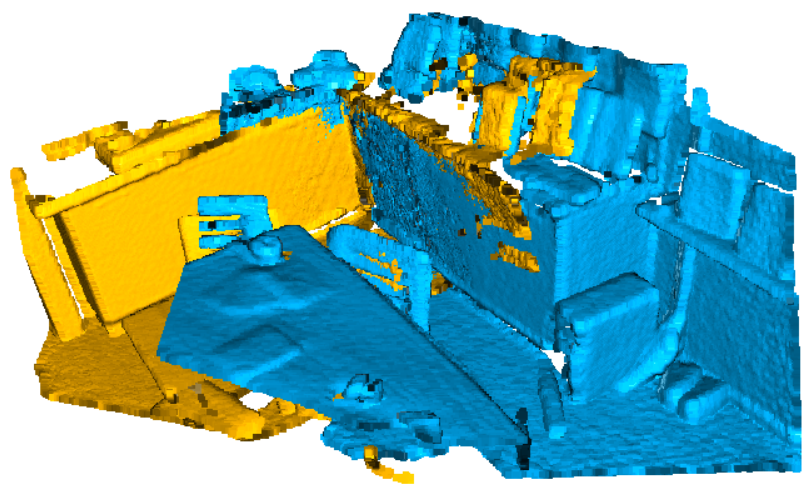}
        \caption*{DHVR}
     \end{subfigure}
     \hspace{0.5em}
\begin{subfigure}[b]{0.26\textwidth}
         \centering
         \includegraphics[width=\textwidth]{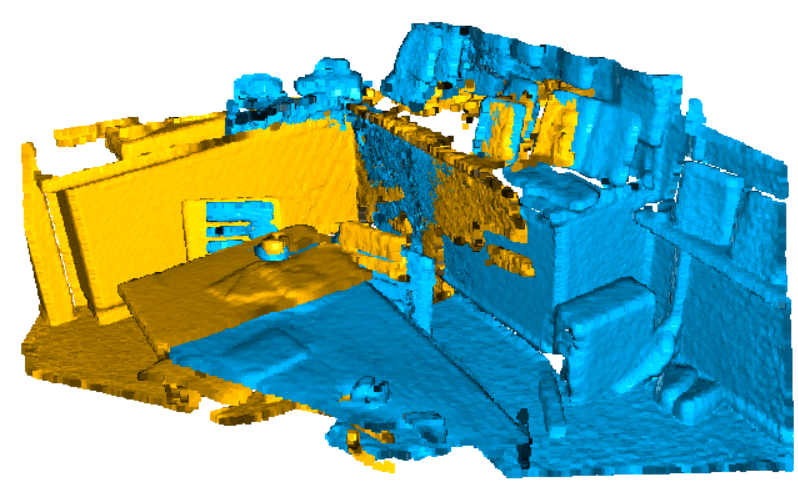}
    \caption*{PointDSC}
     \end{subfigure}
    \hfill
    \begin{subfigure}[b]{0.26\textwidth}
         \centering
        \includegraphics[width=\textwidth]{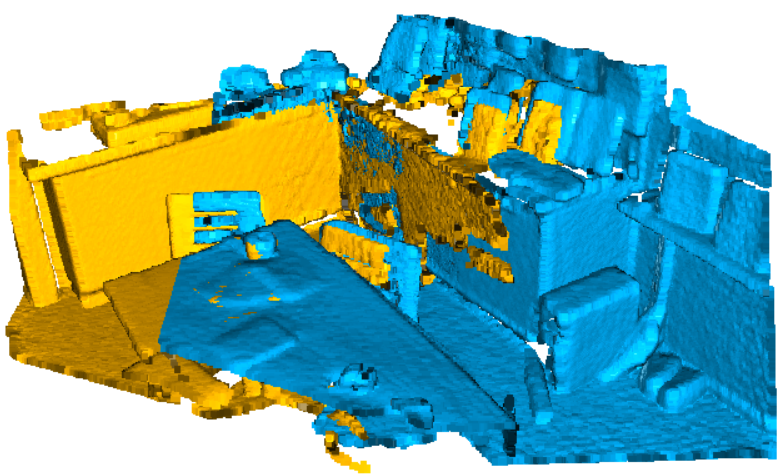}
    \caption*{Ours + DGR}
     \end{subfigure}
     \hspace{0.5em}
      \begin{subfigure}[b]{0.26\textwidth}
         \centering
         \includegraphics[width=\textwidth]{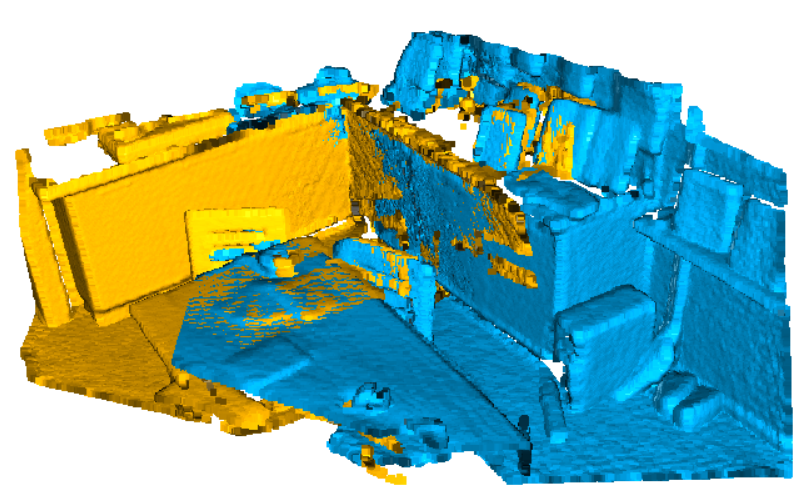}
        \caption*{Ours + DHVR}
     \end{subfigure}
     \hspace{0.5em}
\begin{subfigure}[b]{0.26\textwidth}
         \centering
         \includegraphics[width=\textwidth]{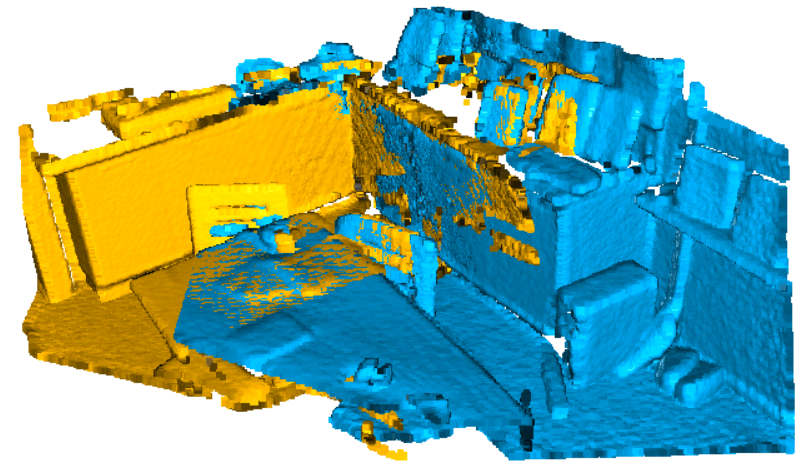}
    \caption*{Ours + PointDSC}
     \end{subfigure}
    \hfill
    
    \begin{subfigure}[b]{0.26\textwidth}
         \centering
        \includegraphics[width=\textwidth]{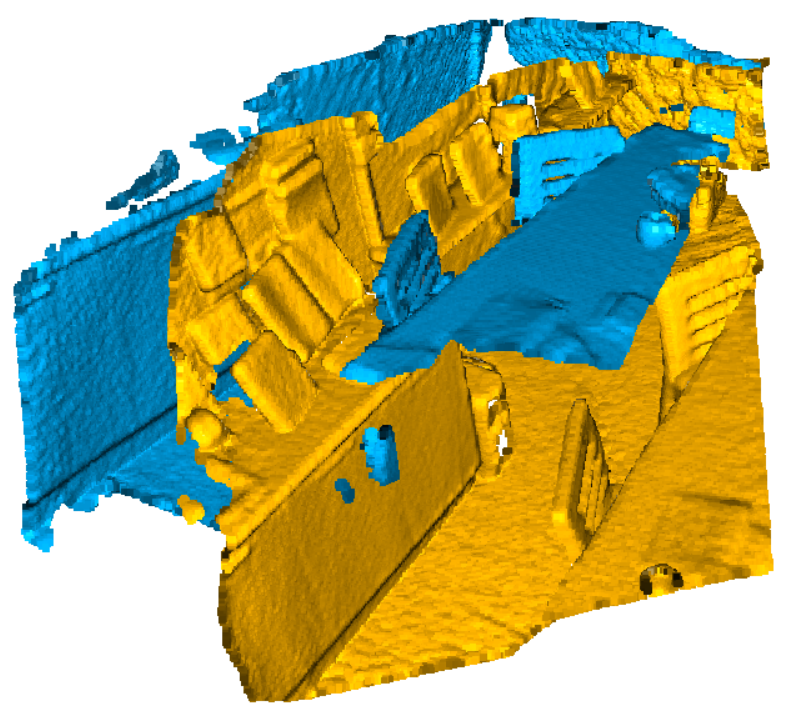}
    \caption*{DGR}
     \end{subfigure}
     \hspace{0.5em}
      \begin{subfigure}[b]{0.26\textwidth}
         \centering
         \includegraphics[width=\textwidth]{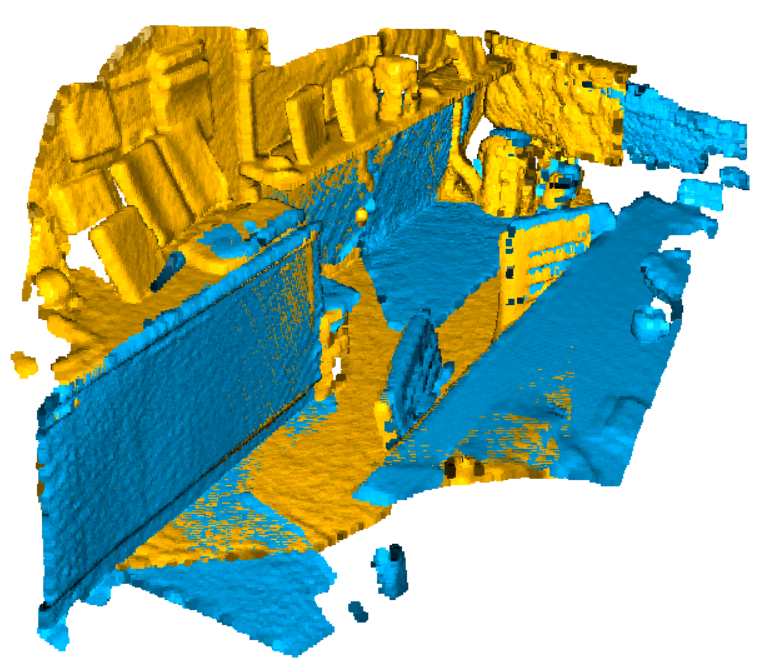}
        \caption*{DHVR}
     \end{subfigure}
     \hspace{0.5em}
\begin{subfigure}[b]{0.26\textwidth}
         \centering
         \includegraphics[width=\textwidth]{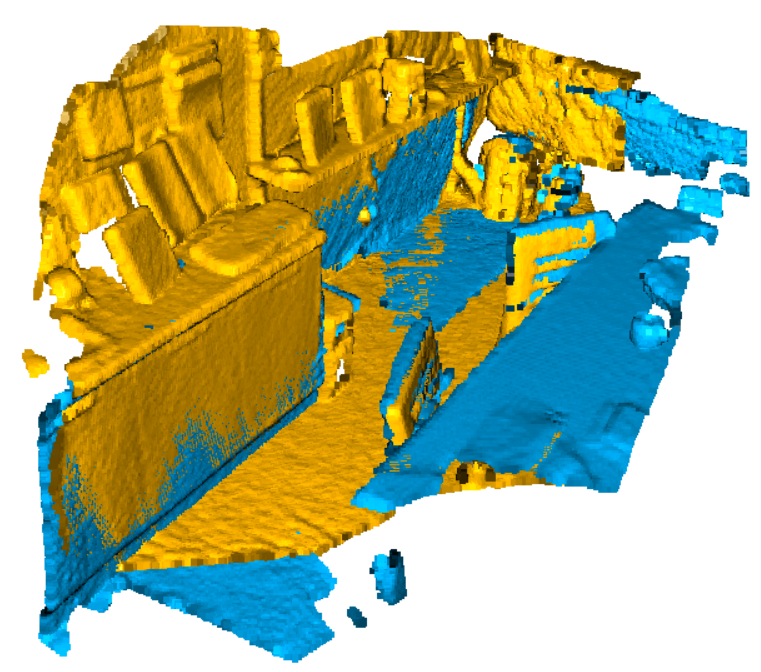}
    \caption*{PointDSC}
     \end{subfigure}
    \hfill
    \begin{subfigure}[b]{0.26\textwidth}
         \centering
        \includegraphics[width=\textwidth]{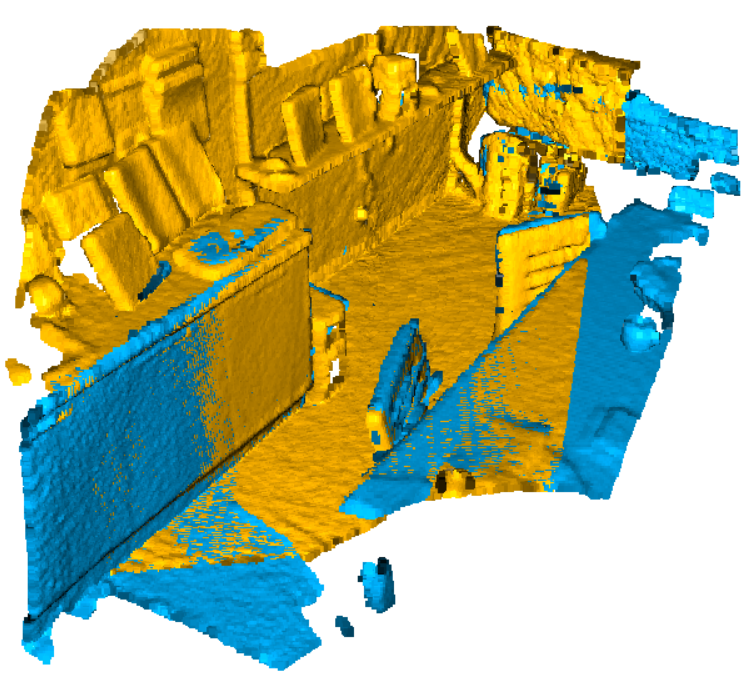}
    \caption*{Ours + DGR}
     \end{subfigure}
     \hspace{0.5em}
      \begin{subfigure}[b]{0.26\textwidth}
         \centering
         \includegraphics[width=\textwidth]{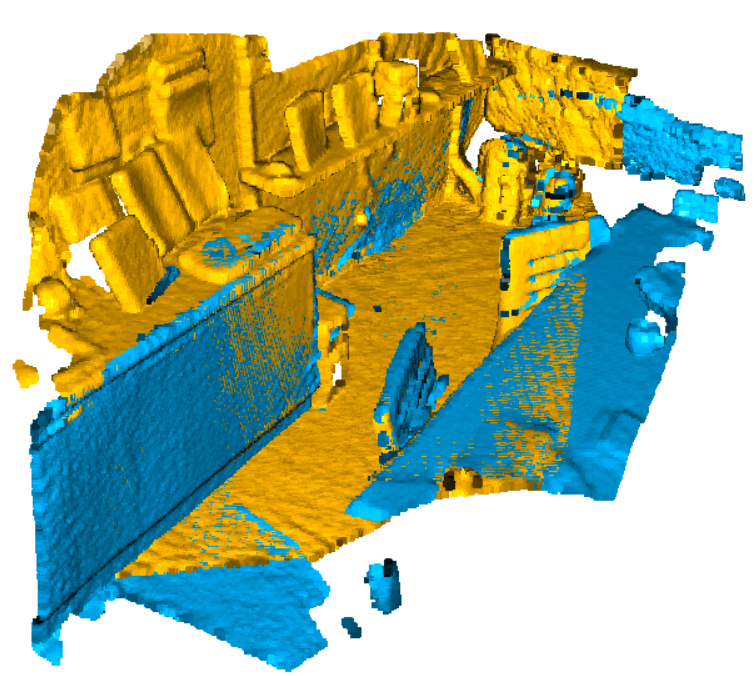}
        \caption*{Ours + DHVR}
     \end{subfigure}
     \hspace{0.5em}
\begin{subfigure}[b]{0.26\textwidth}
         \centering
         \includegraphics[width=\textwidth]{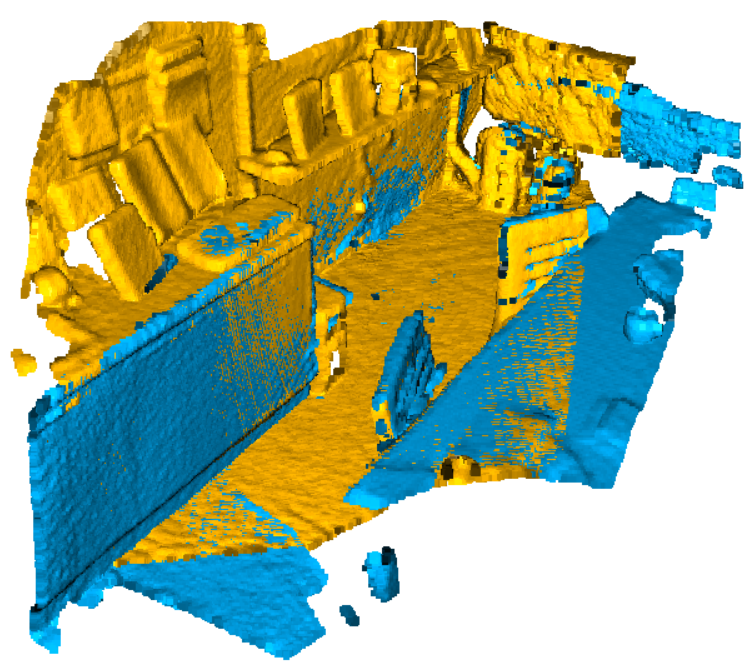}
    \caption*{Ours + PointDSC}
     \end{subfigure}
    \hfill

   \caption{Qualitative results on 3DMatch dataset \cite{3Dmatch}.}
\label{fig:qualitative1}
\end{figure*}

\begin{figure*}
\centering

    \begin{subfigure}[b]{0.26\textwidth}
         \centering
        \includegraphics[width=\textwidth]{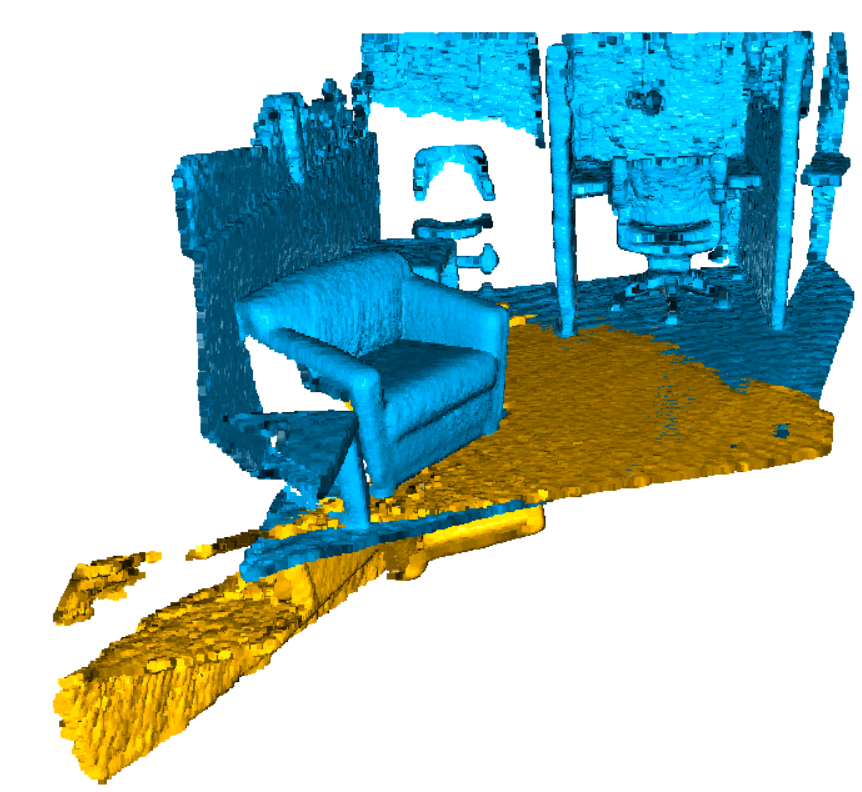}
    \caption*{DGR}
     \end{subfigure}
     \hspace{0.5em}
      \begin{subfigure}[b]{0.26\textwidth}
         \centering
         \includegraphics[width=\textwidth]{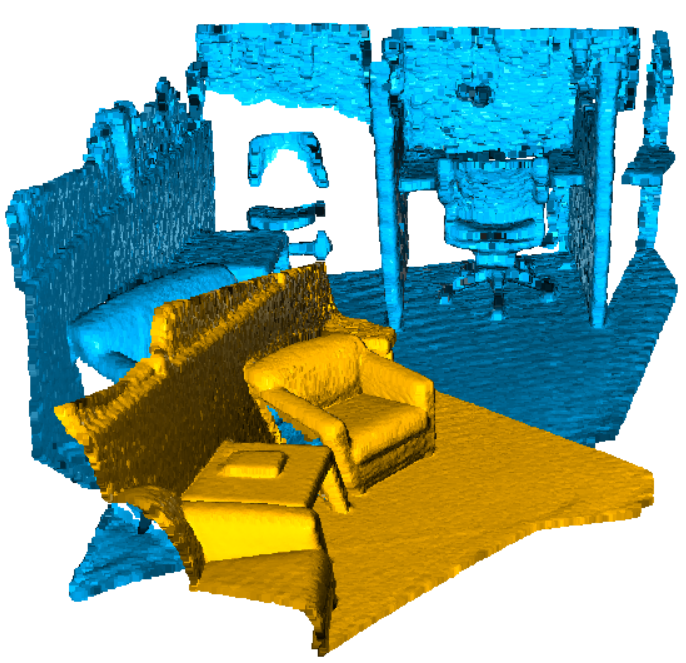}
        \caption*{DHVR}
     \end{subfigure}
     \hspace{0.5em}
\begin{subfigure}[b]{0.26\textwidth}
         \centering
         \includegraphics[width=\textwidth]{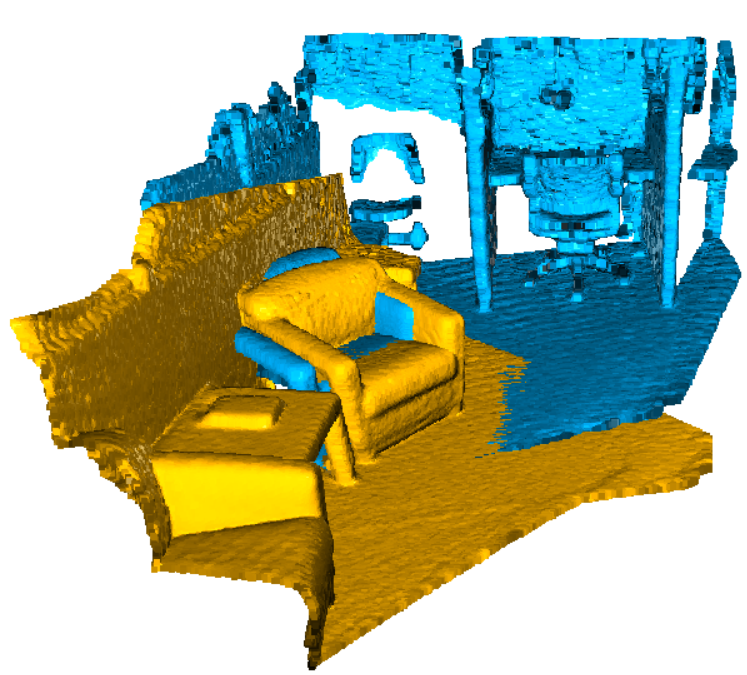}
    \caption*{PointDSC}
     \end{subfigure}
    \hfill
    \begin{subfigure}[b]{0.26\textwidth}
         \centering
        \includegraphics[width=\textwidth]{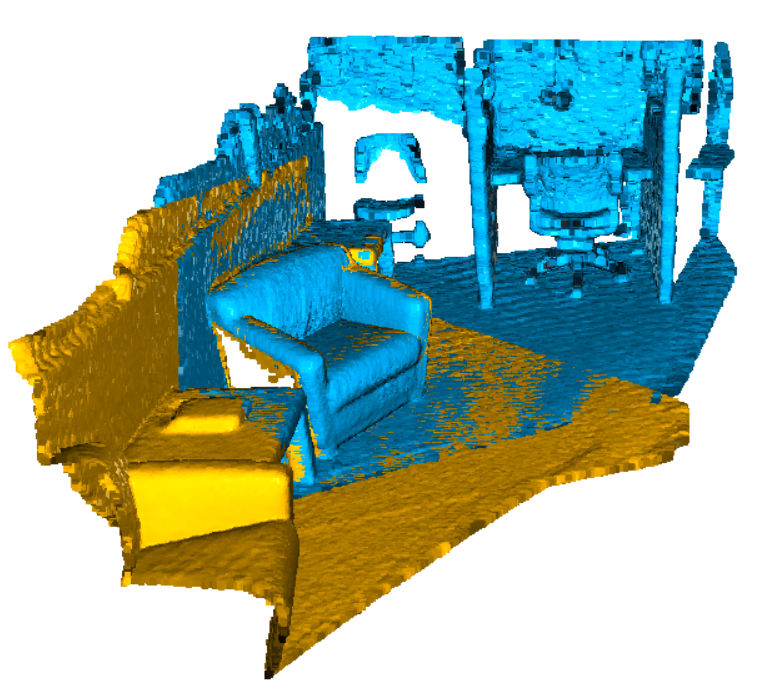}
    \caption*{Ours + DGR}
     \end{subfigure}
     \hspace{0.5em}
      \begin{subfigure}[b]{0.26\textwidth}
         \centering
         \includegraphics[width=\textwidth]{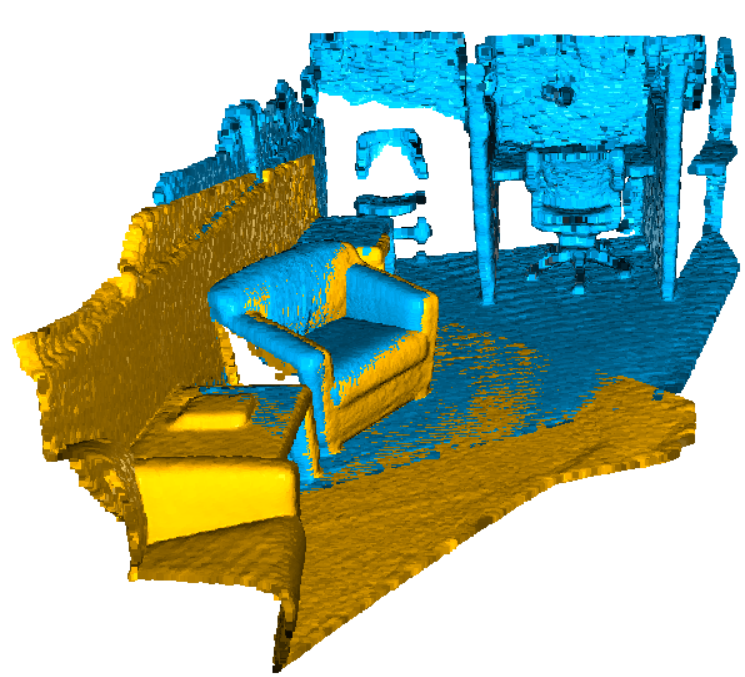}
        \caption*{Ours + DHVR}
     \end{subfigure}
     \hspace{0.5em}
\begin{subfigure}[b]{0.26\textwidth}
         \centering
         \includegraphics[width=\textwidth]{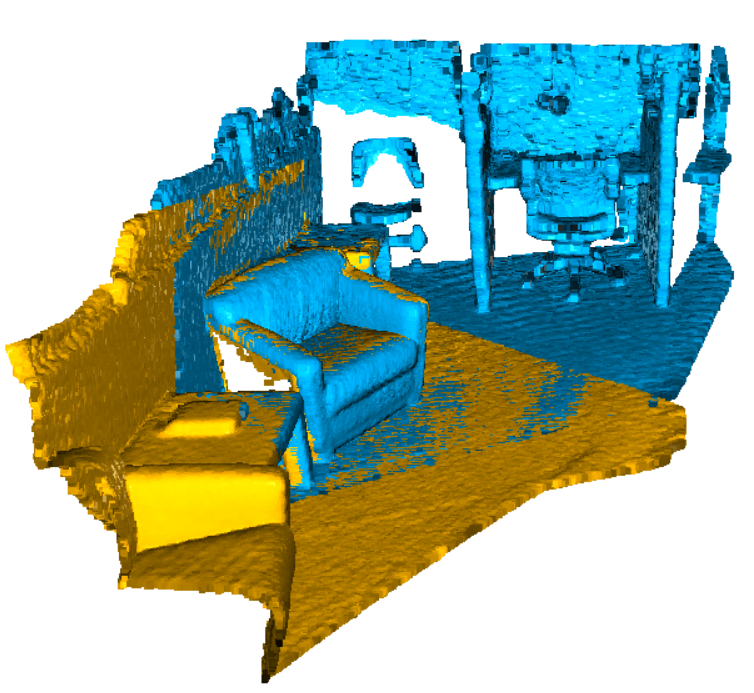}
    \caption*{Ours + PointDSC}
     \end{subfigure}
    \hfill
    \begin{subfigure}[b]{0.26\textwidth}
         \centering
        \includegraphics[width=\textwidth]{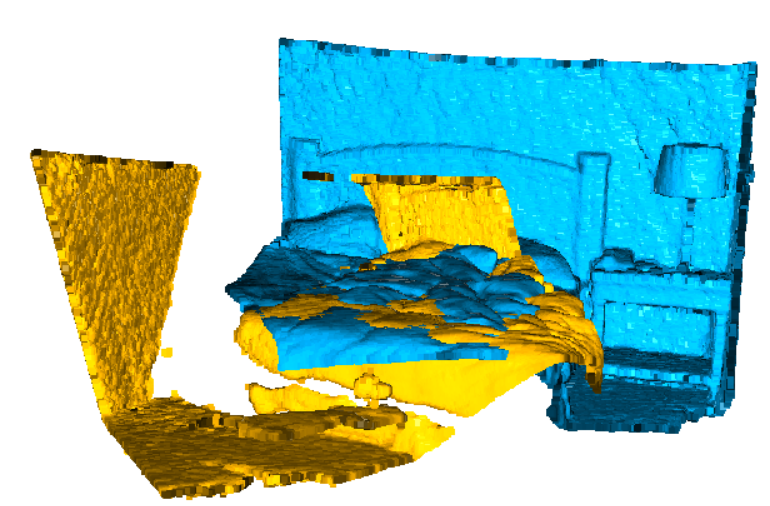}
    \caption*{DGR}
     \end{subfigure}
     \hspace{0.5em}
      \begin{subfigure}[b]{0.26\textwidth}
         \centering
         \includegraphics[width=\textwidth]{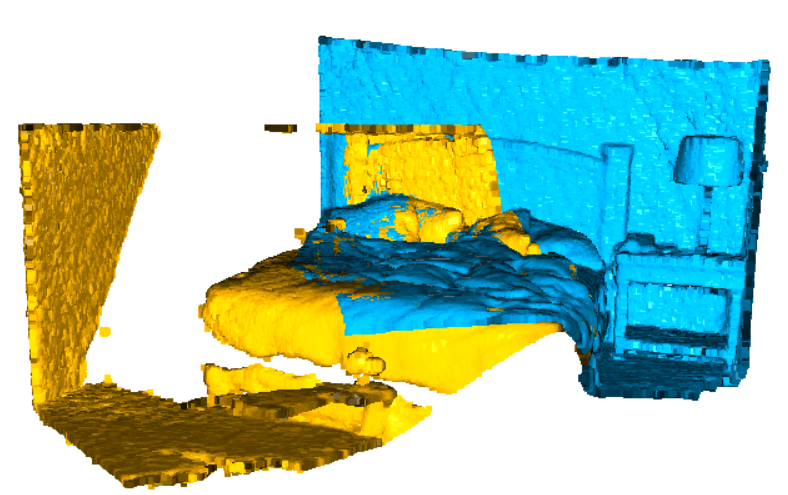}
        \caption*{DHVR}
     \end{subfigure}
     \hspace{0.5em}
\begin{subfigure}[b]{0.26\textwidth}
         \centering
         \includegraphics[width=\textwidth]{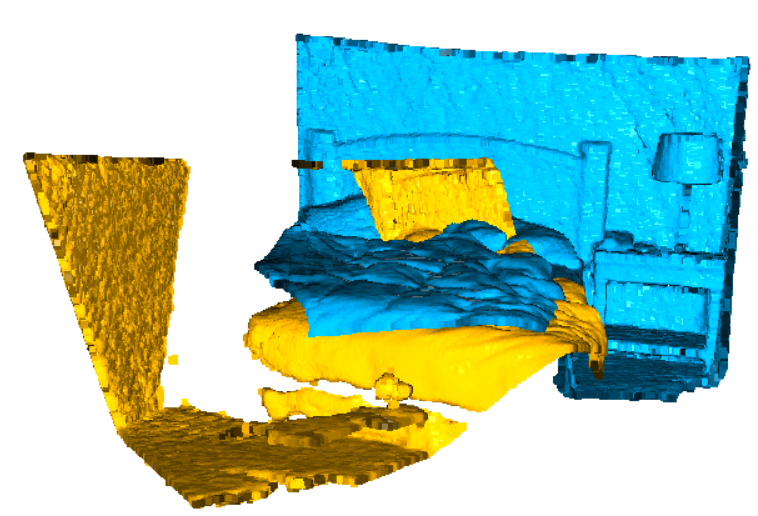}
    \caption*{PointDSC}
     \end{subfigure}
    \hfill
    \begin{subfigure}[b]{0.26\textwidth}
         \centering
        \includegraphics[width=\textwidth]{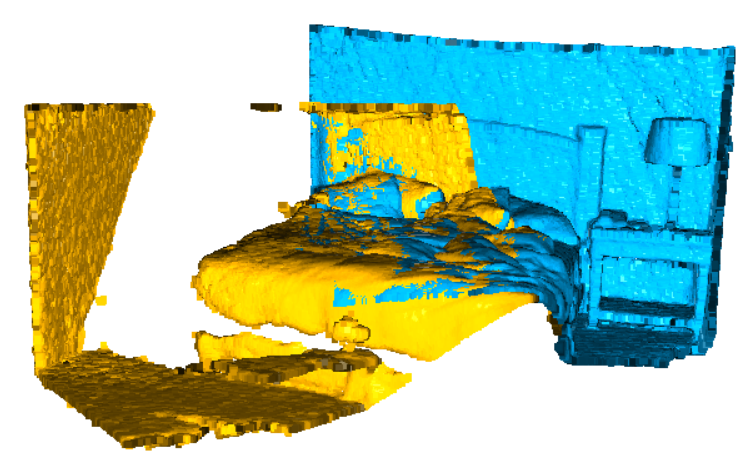}
    \caption*{Ours + DGR}
     \end{subfigure}
     \hspace{0.5em}
      \begin{subfigure}[b]{0.26\textwidth}
         \centering
         \includegraphics[width=\textwidth]{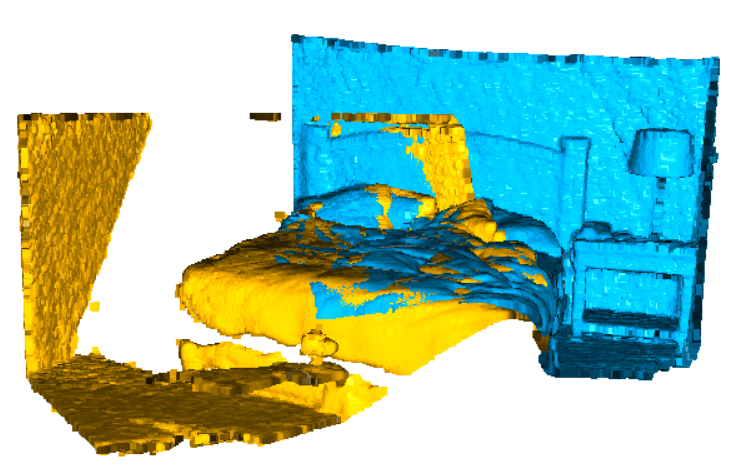}
        \caption*{Ours + DHVR}
     \end{subfigure}
     \hspace{0.5em}
\begin{subfigure}[b]{0.26\textwidth}
         \centering
         \includegraphics[width=\textwidth]{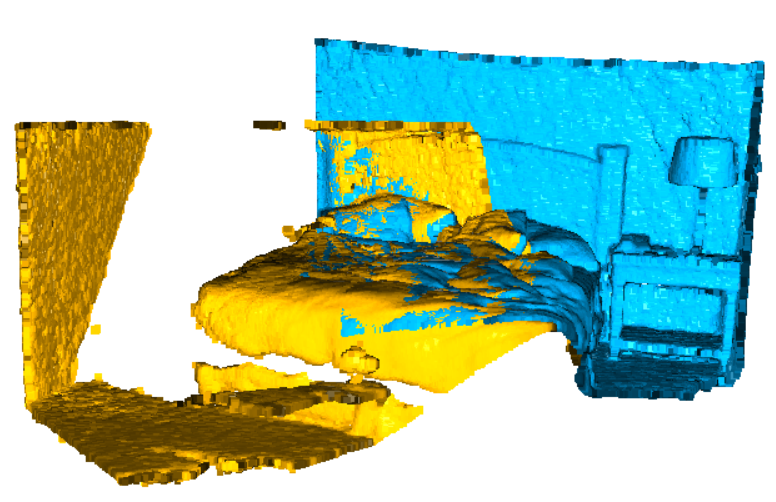}
    \caption*{Ours + PointDSC}
     \end{subfigure}
    \hfill
   \caption{Qualitative results on 3DMatch dataset \cite{3Dmatch}.}
\label{fig:qualitative2}
\end{figure*}

\begin{figure*}
\centering

    \begin{subfigure}[b]{0.32\textwidth}
         \centering
        \includegraphics[width=\textwidth]{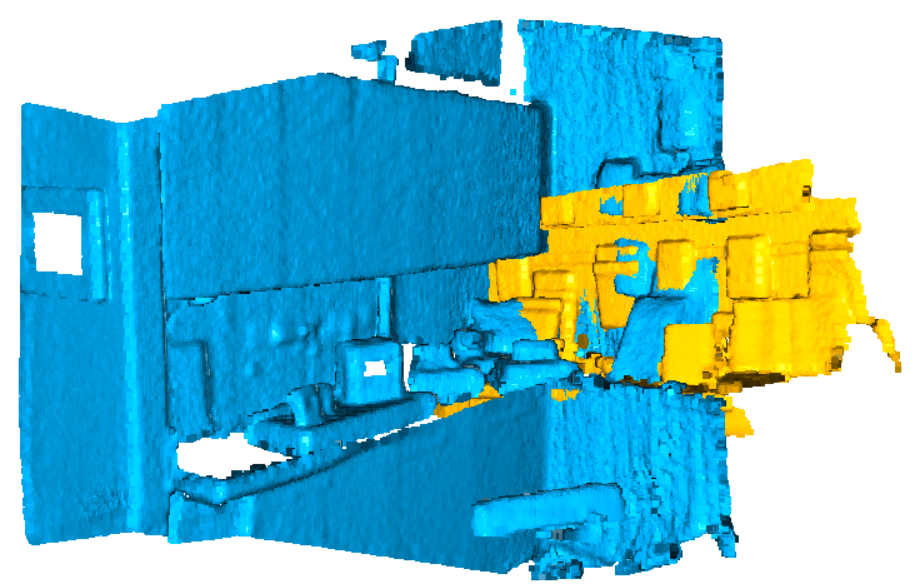}
    \caption*{DGR}
     \end{subfigure}
     \hspace{0.5em}
      \begin{subfigure}[b]{0.3\textwidth}
         \centering
         \includegraphics[width=\textwidth]{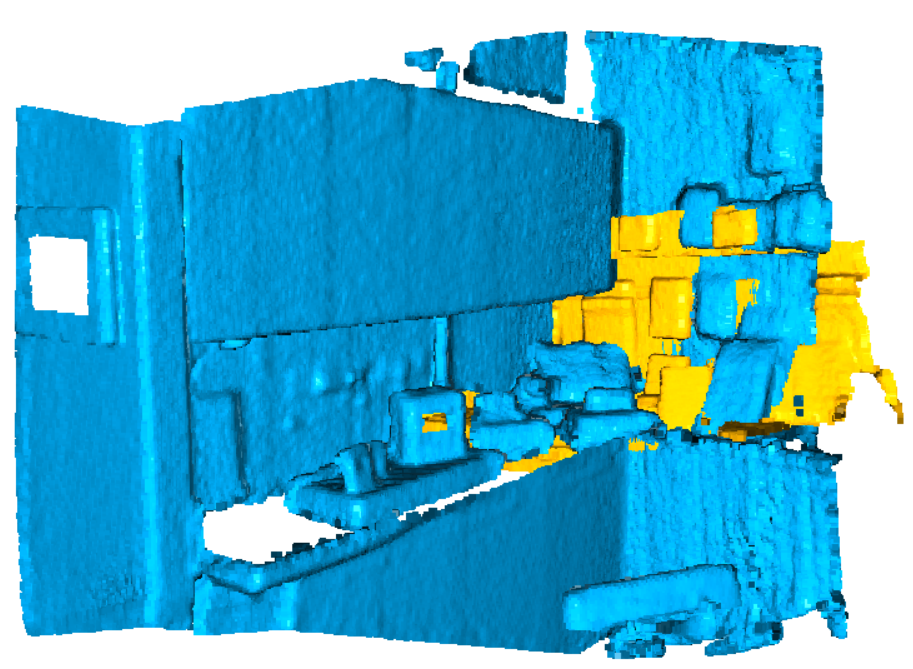}
        \caption*{DHVR}
     \end{subfigure}
     \hspace{0.5em}
\begin{subfigure}[b]{0.31\textwidth}
         \centering
         \includegraphics[width=\textwidth]{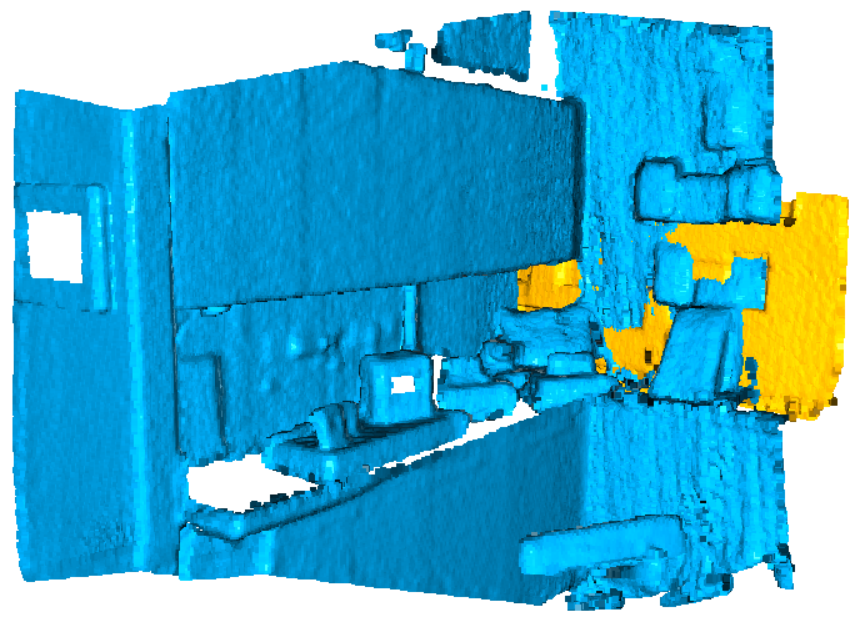}
    \caption*{PointDSC}
     \end{subfigure}
    \hfill
    \begin{subfigure}[b]{0.32\textwidth}
         \centering
        \includegraphics[width=\textwidth]{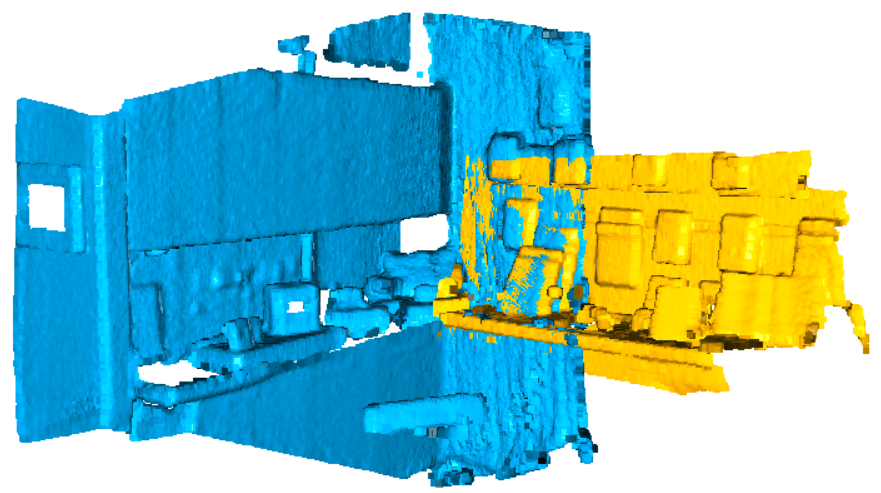}
    \caption*{Ours + DGR}
     \end{subfigure}
     \hspace{0.5em}
      \begin{subfigure}[b]{0.32\textwidth}
         \centering
         \includegraphics[width=\textwidth]{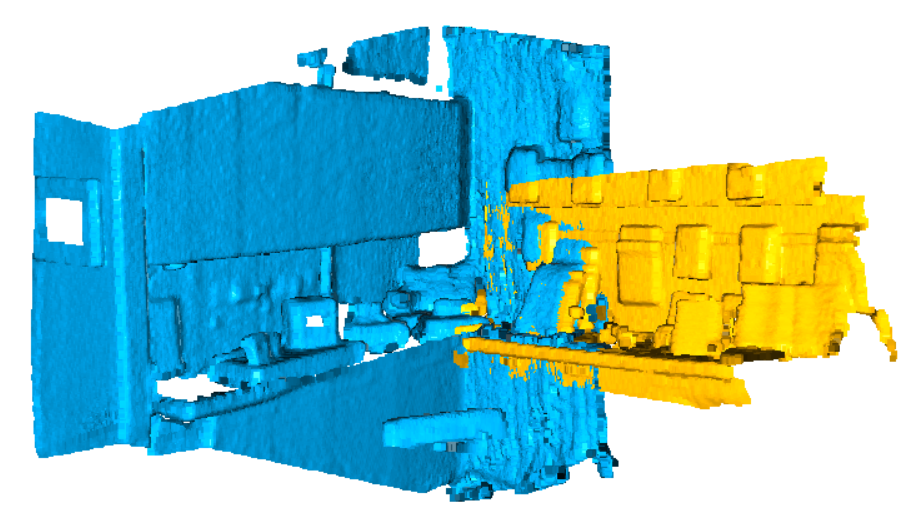}
        \caption*{Ours + DHVR}
     \end{subfigure}
     \hspace{0.5em}
\begin{subfigure}[b]{0.32\textwidth}
         \centering
         \includegraphics[width=\textwidth]{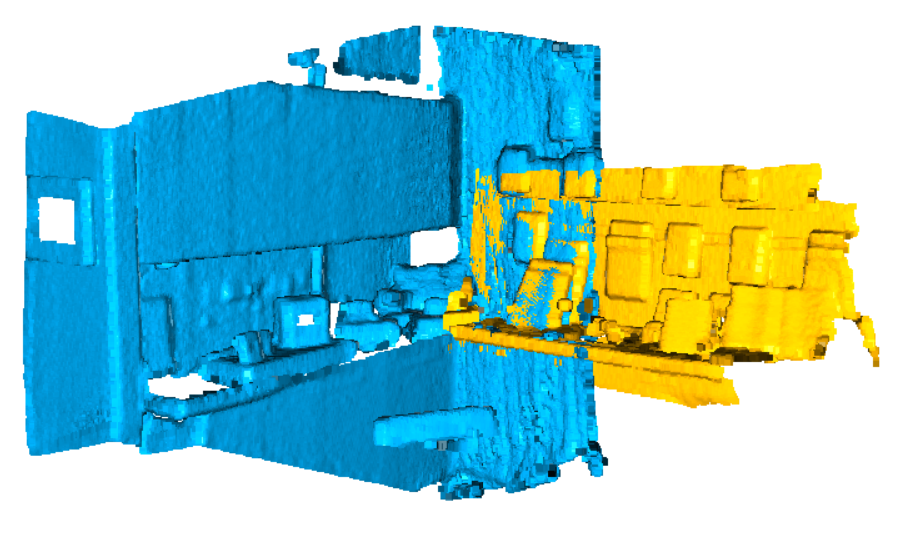}
    \caption*{Ours + PointDSC}
     \end{subfigure}
    \hfill
        \begin{subfigure}[b]{0.32\textwidth}
         \centering
        \includegraphics[width=\textwidth]{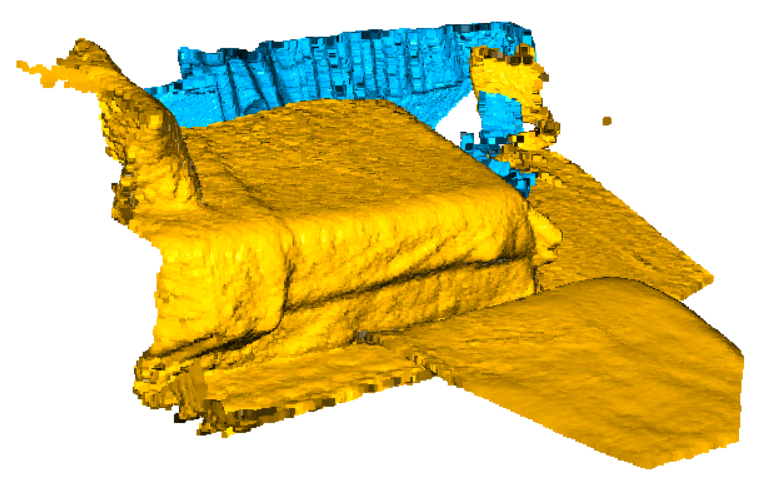}
    \caption*{DGR}
     \end{subfigure}
     \hspace{0.5em}
      \begin{subfigure}[b]{0.3\textwidth}
         \centering
         \includegraphics[width=\textwidth]{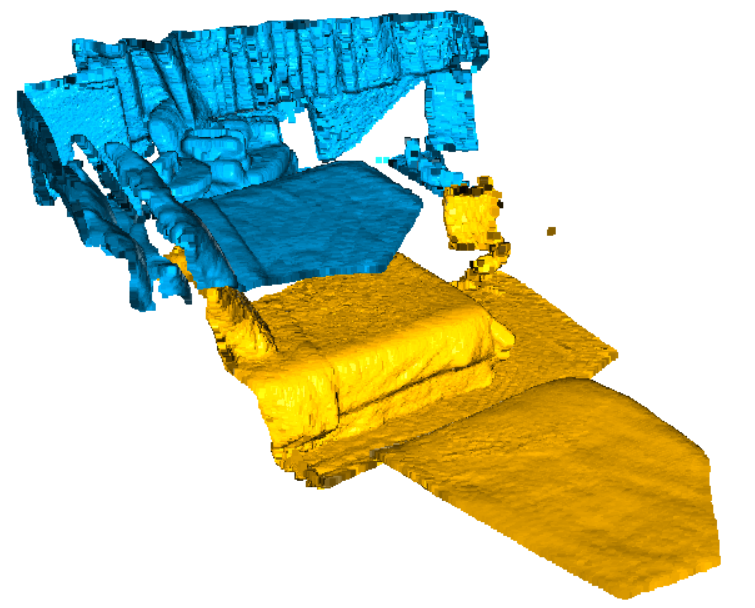}
        \caption*{DHVR}
     \end{subfigure}
     \hspace{0.5em}
\begin{subfigure}[b]{0.31\textwidth}
         \centering
         \includegraphics[width=\textwidth]{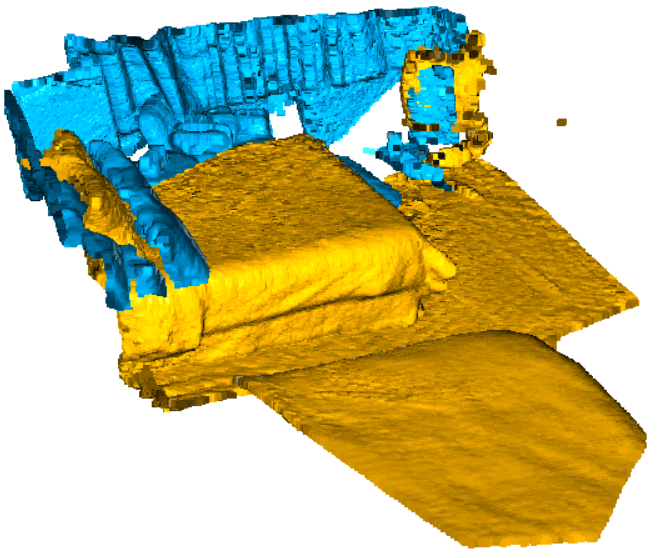}
    \caption*{PointDSC}
     \end{subfigure}
    \hfill
    
    \begin{subfigure}[b]{0.32\textwidth}
         \centering
        \includegraphics[width=\textwidth]{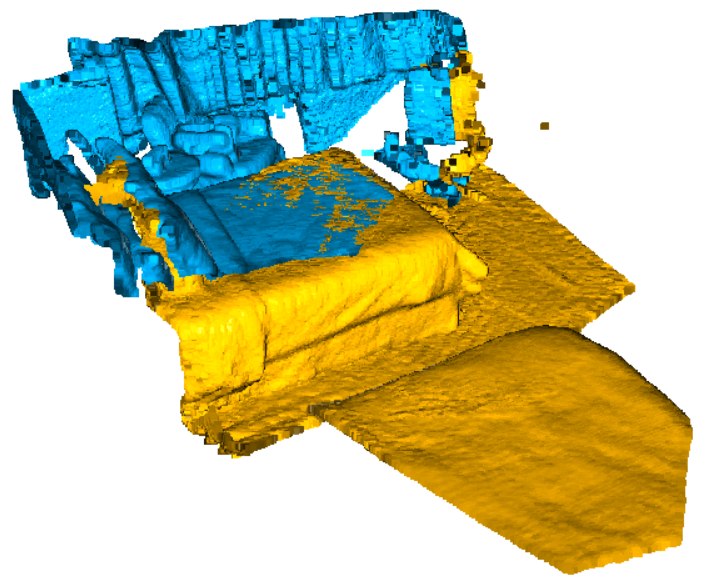}
    \caption*{Ours + DGR}
     \end{subfigure}
     \hspace{0.5em}
      \begin{subfigure}[b]{0.32\textwidth}
         \centering
         \includegraphics[width=\textwidth]{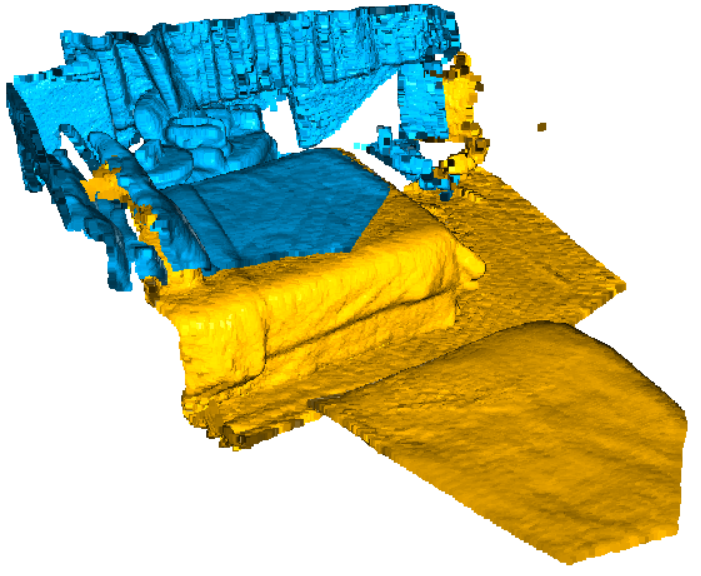}
        \caption*{Ours + DHVR}
     \end{subfigure}
     \hspace{0.5em}
\begin{subfigure}[b]{0.32\textwidth}
         \centering
         \includegraphics[width=\textwidth]{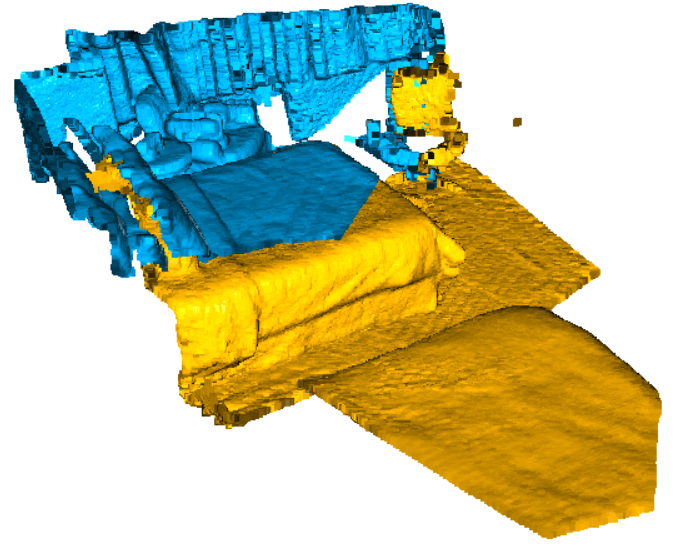}
    \caption*{Ours + PointDSC}
     \end{subfigure}
    \hfill

   \caption{Qualitative results on low-overlapping 3DLoMatch dataset \cite{geoTrans}.}
\label{fig:qualitative_lomatch}
\end{figure*}

\begin{figure*}
\centering
      \begin{subfigure}[b]{0.497\textwidth}
         \centering
        \includegraphics[height=6.5cm]{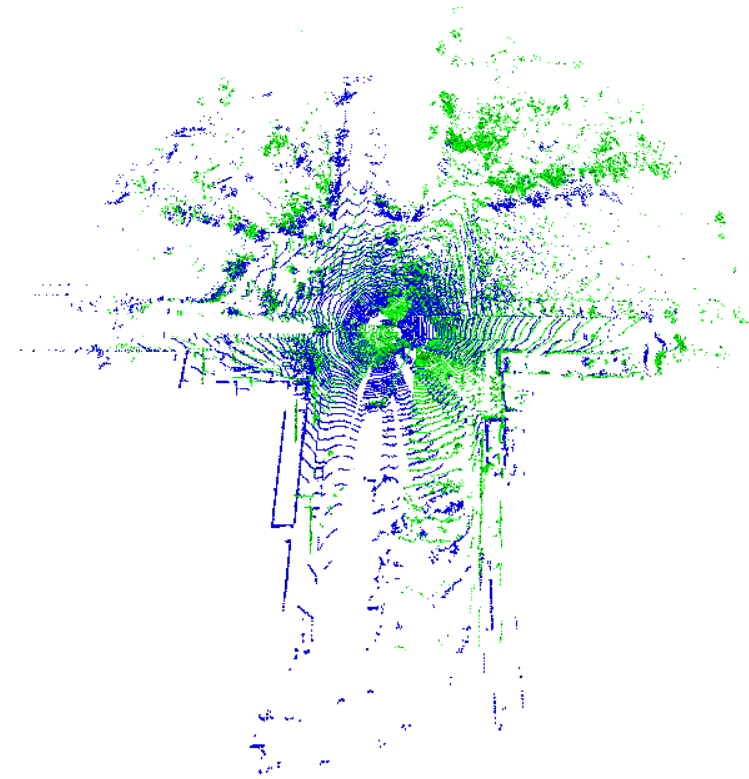}
    \caption*{DGR}
     \end{subfigure}
     \hfill
      \begin{subfigure}[b]{0.497\textwidth}
         \centering
         \includegraphics[height=6.5cm]{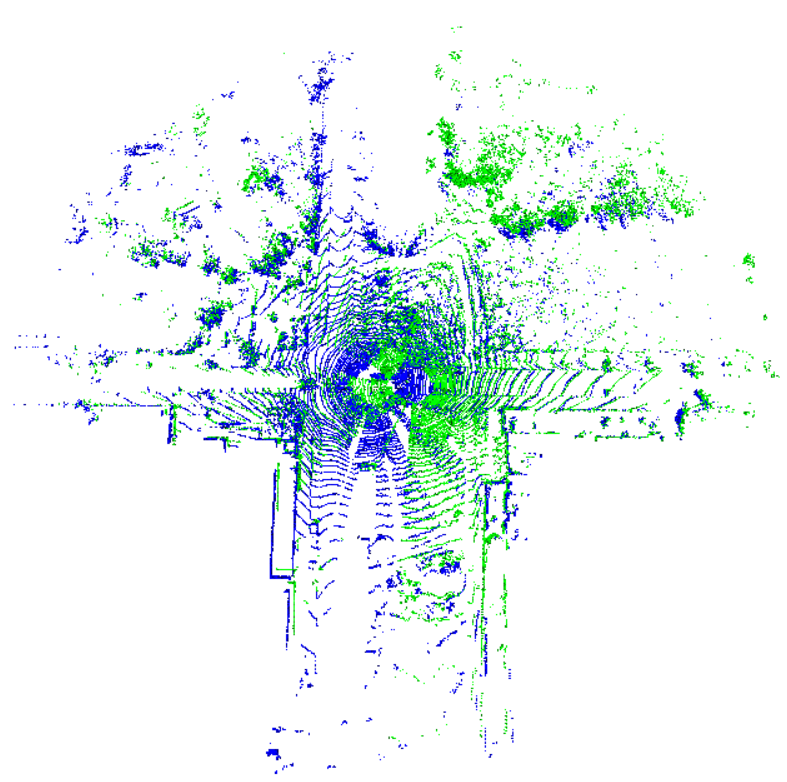}
        \caption*{Ours + DGR}
     \end{subfigure}
     \hfill
\begin{subfigure}[b]{0.495\textwidth}
         \centering
         \includegraphics[height=6.5cm]{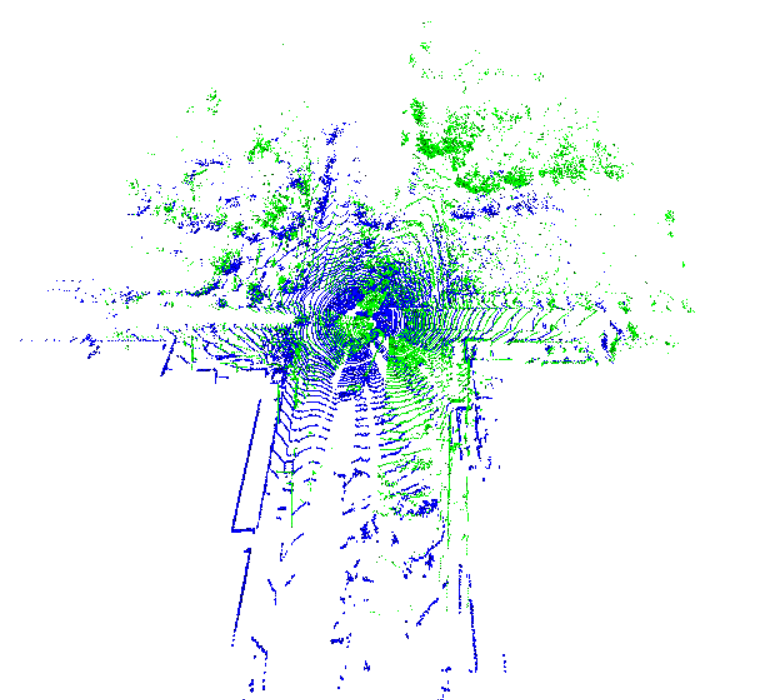}
    \caption*{DHVR}
     \end{subfigure}
    \hfill
    \begin{subfigure}[b]{0.495\textwidth}
         \centering
        \includegraphics[height=6.5cm]{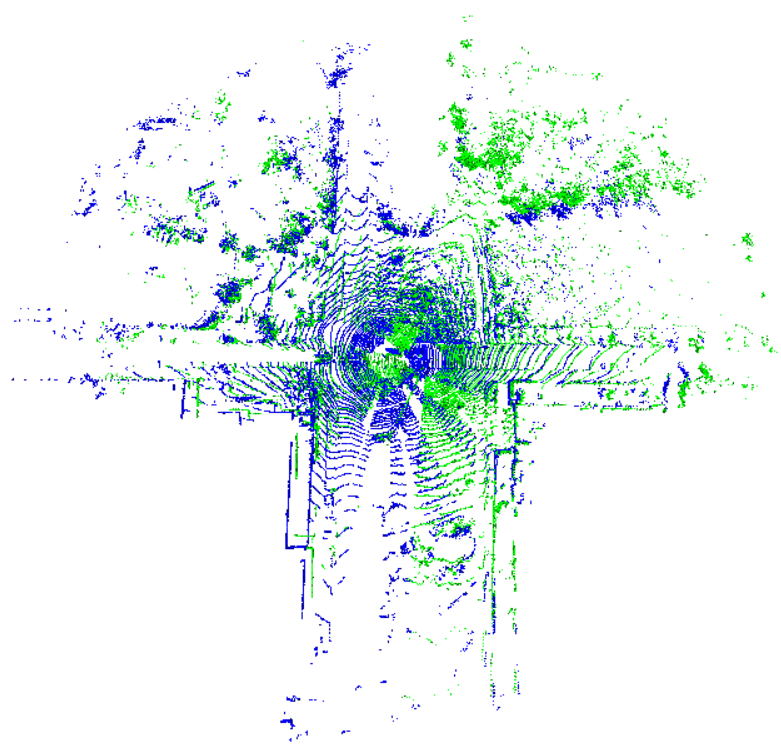}
    \caption*{Ours + DHVR}
     \end{subfigure}
     \hfill
      \begin{subfigure}[b]{0.495\textwidth}
         \centering
         \includegraphics[height=6.5cm]{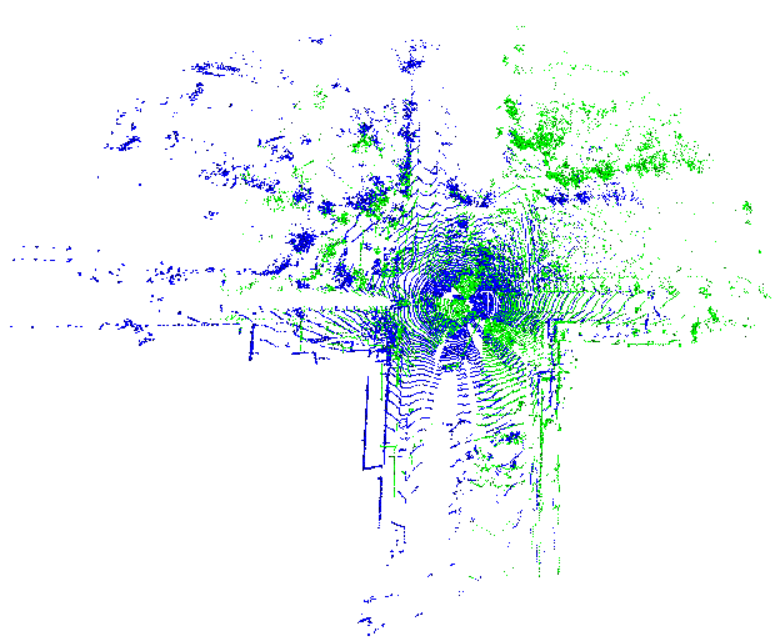}
        \caption*{PointDSC}
     \end{subfigure}
     \hfill
\begin{subfigure}[b]{0.495\textwidth}
         \centering
         \includegraphics[height=6.5cm]{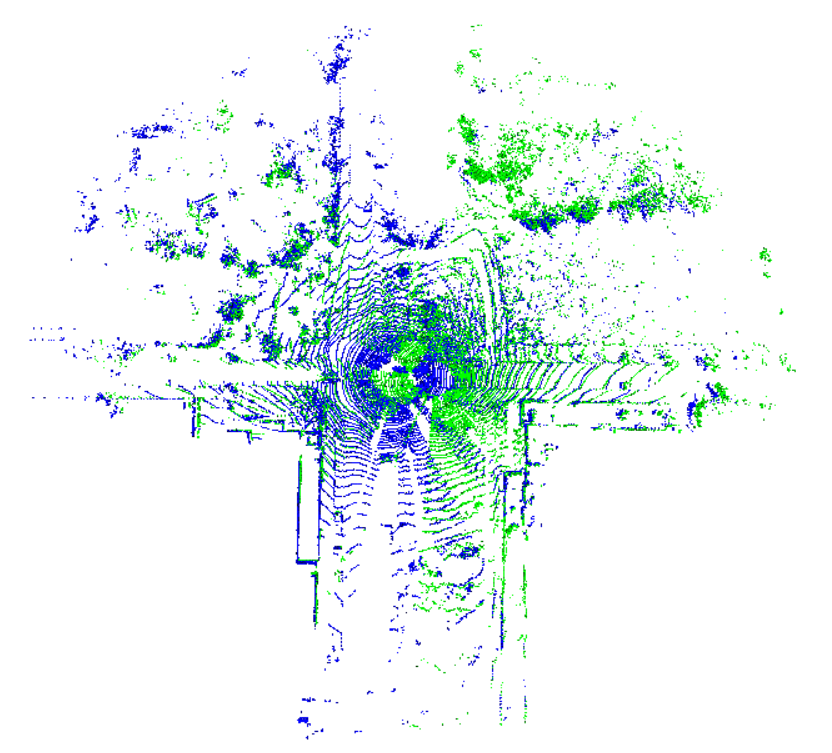}
    \caption*{Ours + PointDSC}
     \end{subfigure}
    \hfill
     \caption{Qualitative results on KITTI dataset \cite{kitti}.}
\label{fig:qualitative_kitti}
\end{figure*}

\begin{figure*}
\centering
      \begin{subfigure}[b]{0.487\textwidth}
         \centering
        \includegraphics[height=6.5cm, width =0.95\textwidth]{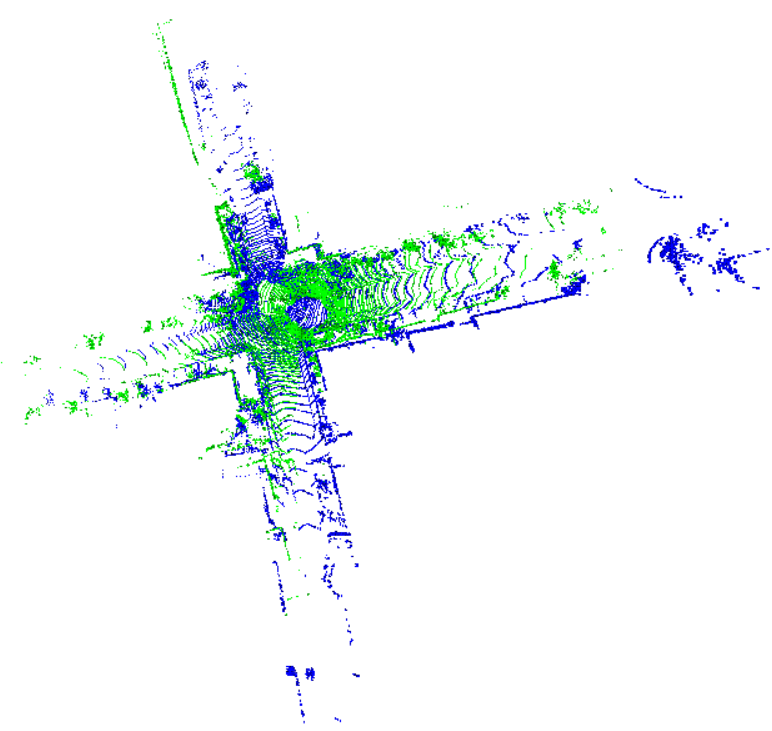}
    \caption*{DGR}
     \end{subfigure}
     \hfill
      \begin{subfigure}[b]{0.487\textwidth}
         \centering
         \includegraphics[height=6.5cm, width =0.95\textwidth]{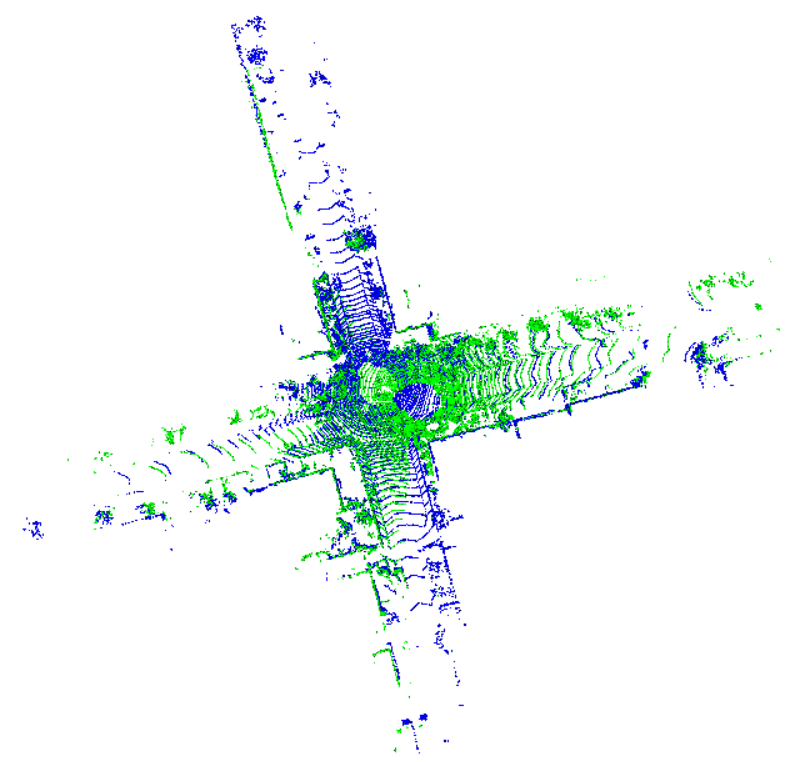}
        \caption*{Ours + DGR}
     \end{subfigure}
     \hfill
\begin{subfigure}[b]{0.487\textwidth}
         \centering
         \includegraphics[height=6.5cm, width =0.95\textwidth]{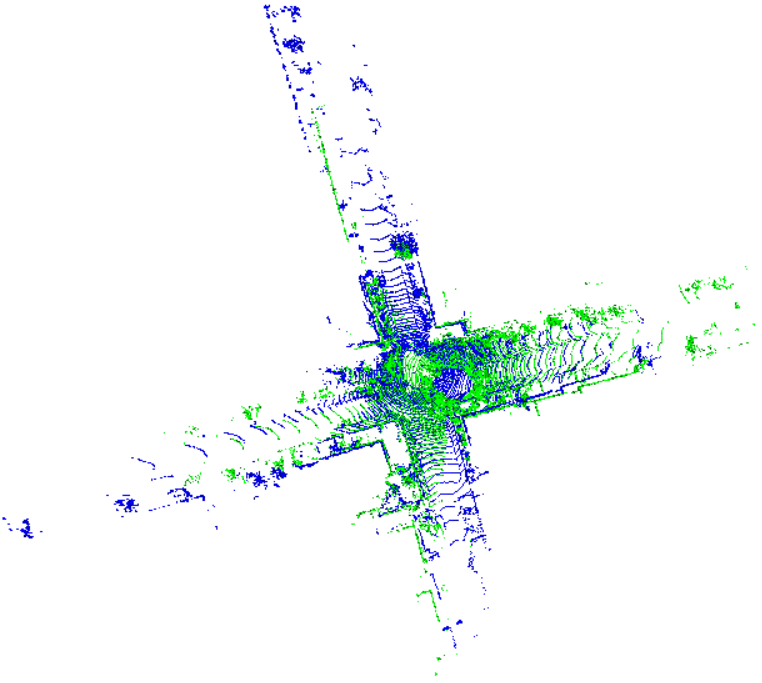}
    \caption*{DHVR}
     \end{subfigure}
    \hfill
    \begin{subfigure}[b]{0.487\textwidth}
         \centering
        \includegraphics[height=6.5cm, width =0.95\textwidth]{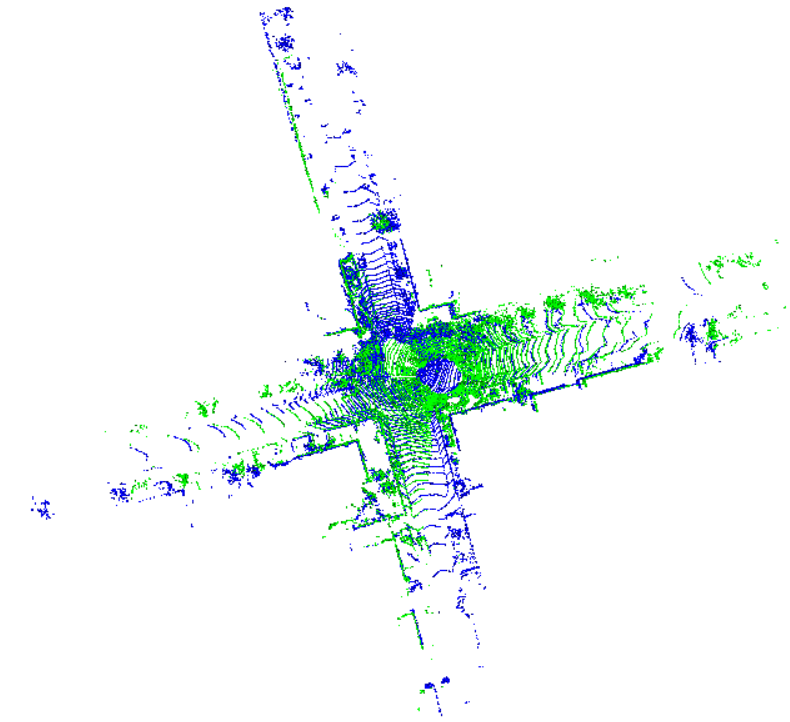}
    \caption*{Ours + DHVR}
     \end{subfigure}
     \hfill
      \begin{subfigure}[b]{0.487\textwidth}
         \centering
         \includegraphics[height=6.5cm, width =0.95\textwidth]{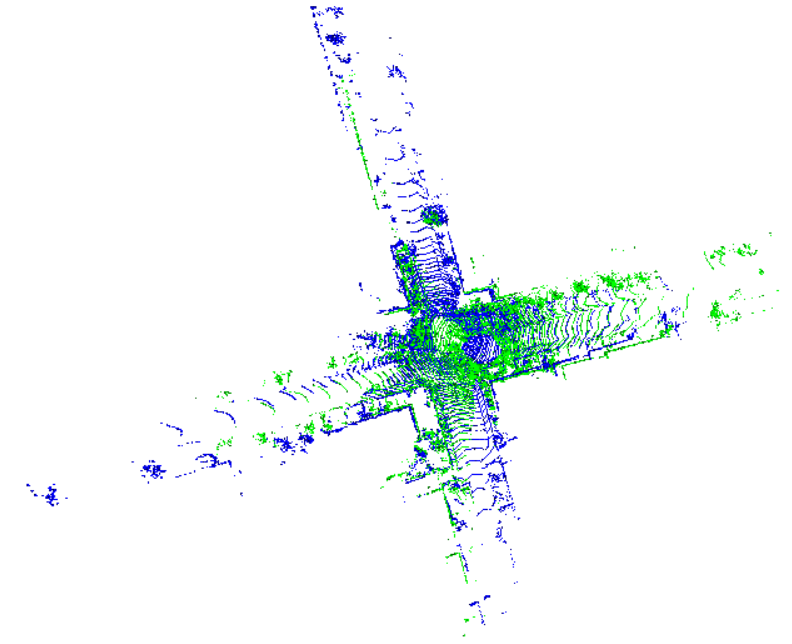}
        \caption*{PointDSC}
     \end{subfigure}
     \hfill
\begin{subfigure}[b]{0.487\textwidth}
         \centering
         \includegraphics[height=6.5cm, width =0.95\textwidth]{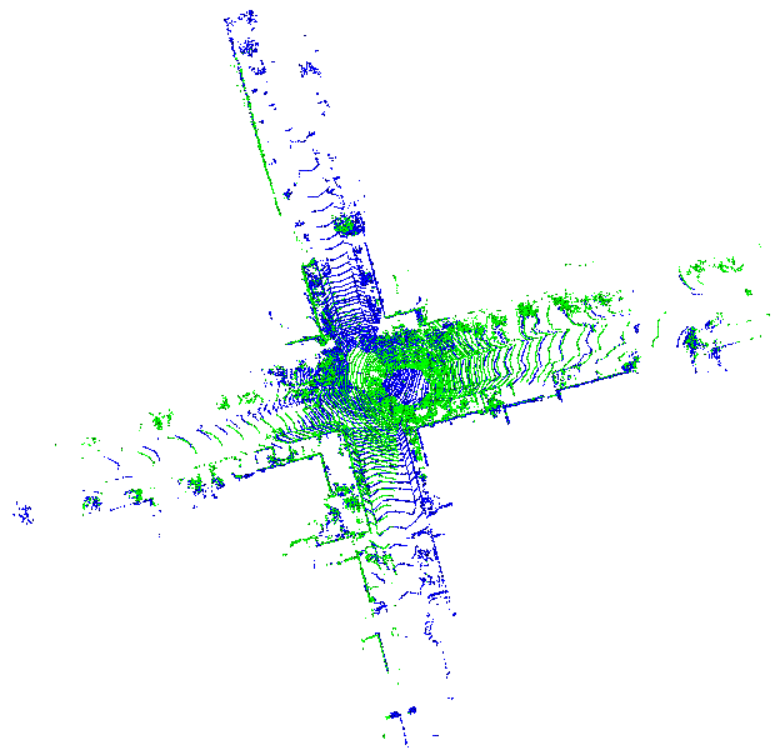}
    \caption*{Ours + PointDSC}
     \end{subfigure}
    \hfill
     \caption{Qualitative results on KITTI dataset \cite{kitti}.}
\label{fig:qualitative2_kitti}
\end{figure*}

We present more qualitative results on 3DMatch (Figure \ref{fig:qualitative1} and Figure \ref{fig:qualitative2}), 3DLoMatch (Figure \ref{fig:qualitative_lomatch}), and KITTI (Figure \ref{fig:qualitative_kitti} and Figure \ref{fig:qualitative2_kitti}).


\end{document}